\documentclass[screen]{acmart}

\usepackage{graphicx}%
\usepackage{multirow}%
\usepackage{amsmath,amsfonts}%
\usepackage{amsthm}%
\usepackage{mathrsfs}%
\usepackage[title]{appendix}%
\usepackage{xcolor}%
\usepackage{textcomp}%
\usepackage{manyfoot}%
\usepackage{booktabs}%
\usepackage{algorithm}%
\usepackage{algorithmicx}%
\usepackage{algpseudocode}%
\usepackage{listings}%
\usepackage{adjustbox}
\usepackage{array}
\usepackage{tabularx}  
\usepackage{booktabs}
\usepackage{xcolor}
\usepackage{soul}
\sethlcolor{yellow}
\usepackage{verbatim}
\usepackage{url}
\usepackage{hyperref}
\usepackage{enumitem}

\setcopyright{othergov}
\copyrightyear{}
\acmYear{2026}
\acmDOI{}
\begin{document}

\title{AI Planning Framework for LLM-Based Web Agents}

\author{Orit Shahnovsky}
\orcid{0009-0002-1587-4504}
\affiliation{%
  \institution{Faculty of Computer and Information Science, University of Haifa}
  \city{Haifa}
  \country{Israel}
}

\author{Rotem Dror}
\email{rdror@is.haifa.ac.il}
\orcid{0000-0002-9433-8410}
\affiliation{%
  \institution{Faculty of Computer and Information Science, University of Haifa}
  \city{Haifa}
  \country{Israel}}

\renewcommand{\shortauthors}{Shahnovsky and Dror}

\begin{abstract}




Developing autonomous agents for web-based tasks is a core challenge in AI. While Large Language Model (LLM) agents can interpret complex user requests, they often operate as black boxes, making it difficult to diagnose why they fail or how they plan.
This paper addresses this gap by formally treating web tasks as sequential decision-making processes. We introduce a taxonomy that maps modern agent architectures to traditional planning paradigms: Step-by-Step agents to Breadth-First Search (BFS), Tree Search agents to Best-First Tree Search, and Full-Plan-in-Advance agents to Depth-First Search (DFS). This framework allows for a principled diagnosis of system failures like context drift and incoherent task decomposition.
To evaluate these behaviors, we propose five novel evaluation metrics that assess trajectory quality beyond simple success rates. We support this analysis with a new dataset of 794 human-labeled trajectories from the WebArena benchmark.
Finally, we validate our evaluation framework by comparing a baseline Step-by-Step agent against a novel Full-Plan-in-Advance implementation. Our results reveal that while the Step-by-Step agent aligns more closely with human gold trajectories (38.41\% overall success), the Full-Plan-in-Advance agent excels in technical measures such as element accuracy (89\%), demonstrating the necessity of our proposed metrics for selecting appropriate agent architectures based on specific application constraints.
\end{abstract}





\maketitle

\section{Introduction}
\label{sec1}

It has been an everlasting challenge of artificial intelligence (AI) research to create autonomous agents: intelligent systems that can perceive their environment, formulate complex plans, and act independently to achieve goals. This ambition finds one of its most complex and dynamic arenas on the web, where agents are required to navigate and solve problems in an ever-changing environment of information and applications designed for human users, not machines. Historically, attempts to automate web-based tasks were brittle, relying on rigid scripts that would break at the slightest change in a website's layout. However, the recent advent of large language models (LLMs) has triggered a paradigm shift. These powerful models, possessing unprecedented capabilities of both language and image processing, are now serving as cognitive engines for a new generation of language-guided agents that can reliably interpret user requests and automate complex computer tasks across the digital world \citep{sodhi2023step,zheng2024gpt,he2024webvoyager,pan2024autonomous,zhang2025webpilot}. 

However, reliably automating these complex web tasks requires more than just a powerful cognitive engine; it demands a coherent strategy. When humans confront an ambiguous or multi-step objective, they naturally decompose it into smaller, manageable sub-tasks and solve them sequentially. For an LLM-based web-agent to emulate this intelligence, it must move beyond simple, reactive responses and engage in deliberate task planning. This planning phase is a crucial intermediary step where the agent determines the optimal sequence of actions required to achieve its pre-defined goal. By incorporating a dedicated planning phase before execution, the agent is encouraged to reason about its options, anticipate challenges, and formulate a robust and efficient path forward. This work, therefore, formally treats complex web-browsing tasks as sequential decision-making processes.

This formal treatment is particularly critical given the current state of AI research. Before the LLM era, agent development was largely dominated by formal planning methodologies, which require explicit world models and action definitions, or methodologies from reinforcement learning (RL), which rely on extensive environmental interaction and reward shaping~\citep{shi2017world, nogueira2016end, liu2018reinforcement, el2012distributed, sirin2004automated}. The new paradigm of LLM-agents often operate as ``black-box'' decision-makers, where the complex planning and reasoning processes are implicitly embedded within the model's underlying model and guided only by natural language. This has led to a rapid proliferation of novel agent architectures (e.g., ~\cite{lee2002intelligent,wang2024survey, shlomov2024grounding, qiu2024llm, mon2025embodied}), but it has also created a significant gap in our understanding. We currently lack a unified framework to analyze how these agents plan, why they succeed or fail, and how they relate to the decades of research that preceded them.

The first major contribution of this paper is to bridge this gap. By leveraging the formalization of web tasks as sequential decision-making processes, \textbf{we introduce a novel mapping between modern LLM-agent architectures and traditional planning methodologies}. This mapping provides the common vocabulary that is currently missing. It serves a dual purpose: first, it allows us to categorize and systematically analyze the diverse, rapidly emerging techniques for building LLM-based agents. Second, by connecting these new methods back to established principles, this framework provides a structured foundation to support the development of more types of web agents.

Beyond mere categorization, we assert that our proposed mapping is critical for diagnosing and analyzing the failures of current LLM-based web agents. For any autonomous system operating in a dynamic environment, failure can generally be traced back to one of two main sources: fundamental flaws in the planning and execution logic guiding the agent, or external noise introduced by LLM hallucination.
Different methodologies taken for planning will determine how well the agent will perform if it is not hallucinating. For example, an agent can, while completing a task, gradually lose sight of its original goal. This phenomenon is referred to as context drift \citep{dongre2025drift}. Over many steps, the original goal simply falls out of memory, replaced by the most recent instructions \citep{han2024llm}. The methodology of the agent determines how well it resists this drift \citep{shlomov2024grounding}. Any algorithm that forces the agent to stick to a plan can help it stay aligned with the original intent even after many steps. 
Another reason for failure is that the agent fails to decompose a task into coherent multi-step actions \citep{koh2024tree}. Consequently, it may lose track of task goals or generate inconsistent plans. 
By exploring the reasons why agents fail, we conclude that many reasons are related to the planning approach the agent employs, and we suggest that choosing an agent based on the need would lead to better performance. 

To confirm these hypotheses and move from diagnosis to demonstrable improvement, a critical component is required: meaningful evaluation metrics. Evaluating the performance of autonomous agents, particularly in complex environments like the web, presents a fundamental and challenging problem in Natural Language Processing (NLP)~\citep{ferrag2025llm}. Since these agents execute tasks through an extended sequence of decisions and actions (a trajectory), their effectiveness cannot be measured solely by the final outcome (e.g., success or failure) as is currently implemented in most evaluation benchmarks for web agents, including WebArena \citep{zhou2023webarena} and MiniWoB \citep{shi2017world}. Instead, effectiveness depends equally on the quality of the process leading to that outcome. Consequently, there is a need for new metrics that assess the efficiency, coherence, and stability of the agent's internal plan, in order of analyzing and mitigating planning-related failures.


The coarse binary evaluation inherently limits insight into agent behavior. For example, if a goal requires listing five reviews and the agent correctly retrieves four, the agent receives no recognition for the substantial progress made; the result is simply deemed a failure. 
In addition, these coarse metrics fail to capture crucial aspects of the execution trajectory, such as the agent's ability to recover from a mistake, follow an efficient path, or avoid repeating unnecessary actions. To address these limitations, the second major contribution of this paper is \textbf{a novel evaluation framework that assesses the entire process of task execution} rather than solely the final outcome. To the best of our knowledge, this work is the first to propose a comprehensive set of metrics designed to quantify the coherence and efficiency of the agent's planning behavior.

To effectively test our novel metrics and empirically demonstrate the utility of our planning taxonomy, we require high-quality ground truth data that capture the entire execution path, not just the final state. Thus, our third contribution is \textbf{the creation of a comprehensive reference-trajectory dataset}. We achieved this by meticulously annotating the widely-used WebArena benchmark with a complete set of human-labeled gold-standard execution traces. This creation of 794 traces based on human performance provides the necessary rich, fine-grained information needed to benchmark both agent failures and planning efficiency across different architectural paradigms. Finally, we apply our full framework by conducting targeted experiments comparing the performance of two agents relying on fundamentally different planning paradigms using our proposed metrics.

In conclusion, we make the following contributions in this paper:
\begin{enumerate}
    \item We establish a nomenclature of strategies for developing web-agent based on three distinct planning paradigms. 
    \item We develop an agent capable of completing web browsing tasks using a plan-ahead strategy.
    \item We construct a trajectory dataset of 794 traces based on human performance, based on the WebArena dataset.
    \item We introduce novel evaluation metrics for web agents that go beyond simple pass/fail criteria, capturing how tasks are performed.
    \item We conduct experiments comparing the performance of two agents relying on different paradigms using the proposed metrics.
\end{enumerate}

\section{Related Work}
\label{sec:related_work}
\subsection{AI Planning}
The ability to generate a plan, i.e., a sequence of actions to transition from an initial state to a desired goal state, is a cornerstone of general AI. Automated planning, or AI planning \citep{ghallab2004automated}, is a branch of AI that concerns the generation of strategies or action sequences, typically for execution by intelligent agents or autonomous robots. Traditionally, this planning was solved by applying symbolic reasoning \citep{garcez2023neurosymbolic}, where the problem is formalized as finding a valid sequence of actions in a state-space graph. 

The state-space model \citep{hamilton1994state} represents the world as a set of possible states, all fully observable and known in advance, where each state captures a configuration of the system at a given time. To move from one state to another, a transition action is necessary (this model assumes that states cannot be changed by actions, i.e., the model is static). A sequence of such transition actions is called a trajectory, and a plan can thus be viewed as a trajectory that connects an initial state and a goal state through valid actions. An agent that performs a plan in such space-state model rely on structured search algorithms, like A* \citep{lavalle2006planning} or GraphPlan \citep{blum1997fast}, or specialized planning languages, such as the Planning Domain Definition Language (PDDL) \citep{haslum2019introduction}, to systematically explore possible paths. 
Therefore, the core principle of automated planning lies in search algorithms whose search space is a subset of the state space. These are generally referred to as state-based search algorithms.

Until recently, the planning and reasoning components of agents were predominantly addressed by symbolic, rule-based models and data-driven methodologies. 
Symbolic agents are based on human-defined rules that determine how they act. These systems are clear and efficient for well-defined tasks, but they lack flexibility in open or dynamic environments. Reinforcement Learning \citep{sutton1998reinforcement} was introduced to address this limitation and allow agents to learn adaptable behaviors. RL is a type of machine learning methodology where an agent learns to make decisions by interacting with the environment. It receives rewards or penalties based on its actions and gradually learns a policy that maximizes long-term expected reward through trial and error.

While RL has achieved impressive results in structured or semi structured environments, many real-world tasks remain partially observable and nondeterministic, where the agent cannot fully perceive the environment or predict the exact outcome of its actions (i.e., calculate the reward correctly). This connects directly to the long-studied problem of planning under uncertainty \citep{pryor1996planning}, which has been extensively explored in the field of AI as it directly relates to real world applications such as autonomous vehicles and web problems. Compared to classical planning, non-classical planning relaxes assumptions of full observability and determinism. At execution time, the true state of the world may be only partially known, and actions may lead to stochastic or branching outcomes. Addressing these complexities requires models that explicitly reason about uncertainty, belief states, and probabilistic transitions.

\subsection{LLM-Based Agents}

The introduction of LLMs catalyzed a fundamental change in agent design, enabling a new paradigm where the LLM serves as the central reasoning and planning unit. LLMs' strong generative and reasoning capabilities have fundamentally changed how AI approaches classical planning and search problems~\citep{lee2002intelligent,wang2024survey, shlomov2024grounding, qiu2024llm, mon2025embodied}. For example, LLMs have been adapted to solve complex reasoning tasks by transforming them into language problems. In the Tree of Thoughts (ToT) framework, \citet{yao2023tree} allow models to explore and evaluate multiple reasoning paths before selecting the best solution. Similarly, LLM+P~\citep{liu2023llm+} focus on complex task planning by integrating LLMs with classical planners (like PDDL solvers) to handle discrete state-spaces with deterministic transitions. Other approaches, such as \citet{xin2025bfs}, adapted search algorithms, such as BFS, into methods for LLM-based automatic theorem-proving.

Currently, almost all state-of-the-art web agents are implemented using LLMs as the core decision-making unit \citep{kuai2025web,abou2025agentic}. Examples like Auto-GPT \citep{yang2023auto} demonstrate the potential of LLMs for multi-step internet exploration via iterative prompting, and frameworks like ReAct (Reasoning and Acting) \citep{yao2022react} are foundational to many web agents and allow the LLM to interleave reasoning traces with executable actions.
These new architectures showed their excellent performance on prominent benchmarks like WebArena \citep{zhou2023webarena} and Mind2Web \citep{deng2023mind2web}. While these benchmarks demonstrate that LLMs function as a powerful alternative to the systems of the past, most works still rely on basic success and failure measures to evaluate the performance of these complex web agents. Consequently, we review current evaluation methodologies in web agent research and discuss the gaps resulting from this limited approach.

\subsection{Evaluation of Web-Agents}
Existing evaluation benchmarks of web agents focus primarily on task completion success rates. Benchmarks such as MiniWoB++ \citep{shi2017world} exemplify this approach. In their setup, an episode is considered successful only when the agent reaches the intended goal state, such as navigating to the correct webpage or submitting the correct form. The performance is then summarized using success rates over all tasks. 
Recent work on developing web agents has shifted toward analyzing the agent’s trajectory to improve its agents' decision-making process. \citet{Li2023} introduced a structured reflection approach, where an agent detects its own errors. For instance, realizing a step was invalid, then attempting to correct it. Similarly, \citet{Song2024} developed an agent that learns from its exploration failures and \citet{pan2024autonomous} provided step-by-step evaluations using a text description of a screenshot to decide on a ``progress toward the goal'' or ``actions that do not contribute to the objective''.
However, these methods focus on improving the agent's behavior rather than measuring its performance or evaluating its given trajectory.


As modern agents are increasingly implemented using LLMs, we propose utilizing LLMs for an additional role in this work: evaluation. In particular, we adopt the LLM-as-a-judge framework to compare model-generated trajectories with human gold trajectories. By serving as judges, LLMs can analyze the agent’s behavior and planned trajectory, helping assess task completion beyond simple success rates.

The use of LLMs-as-judges has previously been explored for evaluating web agents. For example, \citet{xue2025illusion} assessed whether a web task was completed successfully by providing an LLM with the agent’s action trace and key screenshots, and prompting it to output a success/failure judgment. While their method can incorporate intermediate steps, it primarily verifies outcomes and does not explicitly evaluate the quality of the agent’s reasoning or the coherence of its action sequence.

In our setting, each trajectory is decomposed into individual steps written as natural language sentences. We therefore prompt the LLM to decide whether two steps, one from the agent and one from the human reference, are semantically equivalent. In addition, we use the LLM to verify the presence of predefined keywords or phrases within a given sentence (e.g., for partial success evaluation). This use of LLMs for semantic comparison is supported by prior work showing that LLMs are effective at such judgments: \citet{gatto2023text} demonstrate strong LLM performance on semantic similarity tasks, and \citet{gilardi2023chatgpt} argue that LLM-based judging offers key advantages in scalability and consistency, which are particularly important for large datasets. Beyond producing scores or binary decisions, LLM judges can also provide explanations, improving interpretability \citep{huang2023chatgpt}; when needed, they can further be queried for clarification.

Building on the insights gained from analyzing the gaps in existing planning and evaluation frameworks, the subsequent section introduces the first major contribution of this work: a comprehensive taxonomy that systematically categorizes LLM-based web agents according to their underlying traditional planning strategies. For clarity, we will exemplify each methodology within this new nomenclature by detailing its implementation using both existing state-of-the-art web agents and novel architectures developed specifically in this work.

\section{Planning-Based Web Agent}
\label{chap:planning_web_agents}


\subsection{Web Agent Taxonomy Inspired by Traditional Planning}
\label{sec:webAgentTaxonomy}

The web-browsing environment can be thought of as a partially observable Markov decision process (POMDP), where an agent navigates a state space $\mathcal{S}$ via an action space $\mathcal{A}$. At step $t$, the webpage state is $s_t \in \mathcal{S}$ (represented by an accessibility tree). The agent selects an action $a_t \in \mathcal{A}$ (e.g., click, type, scroll) to transition the environment to $s_{t+1}$. The goal is to reach a terminal state $s_G$ that satisfies the user instruction. Within this formalization, current LLM agents do not merely mimic classical search; they implement distinct traversal strategies over the $(\mathcal{S}, \mathcal{A})$ graph:

\begin{enumerate}
    \item \textbf{Step-by-Step Agents (Breadth-Oriented Traversal):}\label{sec:stepBystep} The agent generates an implicit candidate set of immediate actions $\mathcal{A}_{candidate} \subset \mathcal{A}$ valid for the current state $s_t$. It evaluates this subset via the LLM's internal representation, selects a single $a_t\in \mathcal{A}_{candidate}$, observes $s_{t+1}$, and recalculates $\mathcal{A}_{candidate}$. This strictly limits the search horizon to depth $d=1$ at each time step, prioritizing immediate state feedback over long-term horizon planning.
    


    Several existing web agents implement the step-by-step methodology. For example, in the WebArena framework \citep{zhou2023webarena}, the model observes the current webpage, interprets the task, and generates a single action to execute before evaluating again the new state. This design mirrors the breadth-first principle, where every immediate possibility is considered before deepening the search. Likewise, Mind2Web \citep{deng2023mind2web} adopts a similar iterative cycle. 

    \item \textbf{Tree Search Agents (Heuristic-Guided State-Space Traversal):} Unlike step-by-step agents that discard alternative branches after executing an action, Tree Search agents explicitly maintain a search tree $\mathcal{T}$ of explored states to perform multi-step planning. For a given state $s_t$, the agent generates a set of candidate actions $\mathcal{A}_{candidate}(s_t)$ and predicts the subsequent states $s_{t+1}$. To guide the traversal, the agent relies on a pre-defined value function $V: \mathcal{S} \rightarrow [0, 1]$ rather than the implicit LLM encoded knowledge. This function evaluates the promisingness of a branch by scoring the likelihood that a given state $s$ is the terminal goal state $s_G$, where $V(s_G) = 1$. The algorithm iteratively evaluates the frontier of the search tree and explicitly expands the most promising node $s^* = \arg\max_{s \in \text{frontier}(\mathcal{T})} V(s)$. By prioritizing options using this value function, the agent implements a rigorous best-first search over the $(\mathcal{S}, \mathcal{A})$ environment. 


    \citet{koh2024tree} proposes a search algorithm for LLM agents that explicitly performs exploration and multi-step planning in interactive web environments. They implement a best-first search that employs a value function to prioritize options. 
    The function estimates how close the current state is to achieving the goal, where reaching the goal corresponds to a value of 1. The value function is obtained by combining the different reasoning paths generated by the model based on the agent’s observations, producing scores that are sensitive to small differences and helping the agent decide which states to explore first. 

    \item \textbf{Full-Plan-in-Advance Agents (Depth-Oriented Traversal):} The agent generates a complete planned trajectory $\tau = (a_1, a_2, ..., a_n)$ from $s_0$ to $s_G$ prior to execution. During execution, the agent strictly traverses this specific depth-oriented branch of the state-action graph. If $s_t$ diverges from the expected intermediate state in $\tau$, the execution either fails or requires a complete replanning phase.
    
    
    We claim that an agent planning ahead in this manner conceptually performs a DFS search, as its actions are derived from a complete, pre-calculated plan rather than a single-step decision. Notably, to the best of our knowledge, there are no established web agent implementations that strictly adhere to this pure Full-Plan-in-Advance methodology, prompting us to provide an implementation of this strategy.
    \end{enumerate}

\subsection{Implementation of Full-Plan-in-Advance Agents}
\label{sec:agentImplementation}

Agents that base their next action solely on their current observation often suffer from a significant drawback---they lack foresight. If an agent takes a locally optimal but globally sub-optimal action, it may enter a bad state, which is a state from which it is hard to recover, or it may lead to an inefficient trajectory with unnecessary steps. The Full-Plan-in-Advance approach addresses this gap by requiring the LLM-agent to first generate a complete, multi-step plan of execution before taking any action. This initial plan acts as a global constraint and a clear road map of actions, drawing upon the LLM’s vast encoded knowledge base about general web procedures. 
In our implementation, the agent follows this plan as its primary guiding framework. Crucially, the plan itself is provided to the agent in the prompt for every single reasoning step. This continuous reminder serves as a form of external, high-level memory, helping the agent to interpret new observations, decide how to continue, and strongly resist context drift.

\paragraph{Webpage Representation}
For the agent to perceive its environment and execute steps according to its plan, it requires a structured representation of the webpage. In our implementation, the LLM receives the current state of the webpage as an Accessibility Tree. This tree is a simplified, functional subset of the full Document Object Model (DOM) tree, containing only the elements that are relevant for user interaction and content display. This representation effectively filters out visual noise and non-essential HTML structure, providing the LLM with a clean, navigable input. Every element is represented as its role (e.g., a link, a button, a text input field), its text content (i.e., the user-visible text associated with the element), and its properties (e.g., if the element is focusable, its current value, or other attributes necessary for a complete understanding of its interactive state). 


\paragraph{Plan Generation and Execution} To generate a plan, the LLM agent receives the user’s intent, the accessibility tree of the initial page, and the page’s URL. The agent is instructed to produce a numbered plan, where each step is accompanied by a short descriptive explanation. The actions described in the plan are limited to web based operations such as click, type, or navigate. An example of a generated plan appears in Table~\ref{tab:82HumanVSgenralPlan}. The full prompt can be found in Appendix~\ref{app:promptGeneratePlan}.
Once the plan is generated, it accompanies the agent throughout the execution process, where at each step, in addition to the generated plan, the agent is provided with the current accessibility tree, the active URL, and the task objective. The prompt for each step appears in Appendix~\ref{app:promptPlanAgent}. Based on this input, the LLM generates the next action from an action space that emulates the keyboard and mouse operations commonly available on web pages. Each element is assigned a unique identifier during traversal of the accessibility tree. For example, in Figure~\ref{fig:webarenaStepExample} the action \textit{type [552] [Hobart Street, Pittsburgh] where [522] is [522] textbox `From' required: False} triggers typing into the textbox labeled \textit{From}, which is not mandatory (usually marked in *). The command is then passed to the execution layer, implemented using Playwright\footnote{\url{https://playwright.dev/}}, which performs the actual browser operation.
In Section~\ref{sec:results}, we present the performance of the Full-Plan-in-Advance agent demonstrated on the WebArena benchmark, using the evaluation metrics we introduce in the next section.

\section{Evaluation Metrics}
\label{sec:evaluationMetrics}

As previously mentioned, the evaluation of web agents relies mostly on binary success–failure metrics calculated across a series of tasks. Structurally, evaluation benchmarks for web agents are designed as interactive environments paired with a dataset of natural language instructions and defined success criteria. A prime example is the WebArena benchmark \citep{zhou2023webarena}, which assesses performance by calculating the percentage of tasks that an agent successfully executes within fully functional, self-hosted web applications.

To better understand the agent’s behavior and planning capabilities, we introduce a suite of evaluation metrics that goes beyond a simple pass/fail criteria. This set consists of five metrics designed to measure the quality of the agent's planned and executed trajectories in addition to the binary task completion indicator. 
This distinction is crucial: while a web agent may fail a specific task, its underlying ability to generate a high-quality, logical sequence of actions can indicate a strong capacity for success when faced with similar challenges in real-world deployment, i.e., to test its generalizability. 
Consequently, failing a task in the benchmark does not necessarily mean the agent is of low quality; it may indicate that the agent fulfilled the task partially (e.g., encountered an unexpected pop-up), or that the scenario presented in the benchmark was uniquely challenging, rare, or contained uncommon UI (user interface) elements.

Some of the developed metrics compare the trajectory planned and executed by the agent to the steps taken by a human for completing the same task, i.e., they are reference-based metrics that rely on a given gold standard. For this reason, we created a new annotated dataset based on the tasks in WebArena. The introduction of this dataset and the annotation process appear in Section~\ref{sec:dataset}. The definitions of the proposed metrics are as follows:

\begin{definition}[\textbf{Recovery Rate}]
Measures how well an agent can return to the expected human-demonstrated sequence of actions after deviating from it. A \textbf{deviation incident} is recorded when an agent step does not directly fulfill the current human gold step. A \textbf{recovery} occurs when the agent performs an action that successfully fulfills a subsequent human gold step. 
\[
\textbf{Recovery Rate} =
\frac{1}{\# \text{tasks}}\sum_{t \in \text{tasks}}\frac{\text{\# recoveries in task $t$}}{\text{\# deviation incidents in task $t$}}
\]
\end{definition}

To detect recoveries, the algorithm examines the agent’s subsequent actions and checks whether any of them match one of the upcoming human gold steps. The number of future human steps that may be considered is controlled by a user-defined parameter provided when running the metric. 
Semantic matching between human and agent actions (up to a maximum sequence length of five steps) is performed by an LLM. A predefined semantic similarity score, where 1 means identical and 0 means completely different, determines the minimum score required for the LLM to classify two actions as a semantic match.

In Table~\ref{tab:HumanVsAgentTrajectory}, the agent recovers twice from deviations. First, it deviates by clicking the `About Us' link (A1) instead of the `Products' link (H1). Second, it deviates by scrolling down (A4) instead of clicking the `Electronics' category (H2). Despite these two deviations, the agent eventually performs the correct actions (A2, A3), resulting in a Recovery Rate of $2/2 = 1.0$.

\begin{table*}[h]
\centering
\caption{Human and agent trajectories for the task: Show me all smartphones.}
\begin{tabularx}{\textwidth}{|cX|cX|cX|}
    \hline
    \multicolumn{2}{|c|}{\textbf{Human Trajectory}} & \multicolumn{2}{|c|}{\textbf{Agent Planned Trajectory}}  & 
    \multicolumn{2}{|c|}{\textbf{Agent Actual Trajectory}}\\
    \hline
    Step No. & Action & Step No. & Action & Step No. & Action  \\
    \hline
    H1 & Click Products link & AP1 & Click About Us link & AA1 & Click About Us link\\
    H2 & Click Electronics category & AP2 & Click Products link & AA2 & Click Products link \\
    H3 & Click Smartphones filter & AP3 & Click Electronics category & AA3 & Click Electronics category \\
       &                            & AP4 & Scroll down page           & AA4 & Scroll down page \\
       &                            & AP5 & Click Smartphones filter & AA5 & Click Smartphones filter\\
       &                            & AP6 & Click Smartphones filter & AA6 & Click Back \\
    \hline
\end{tabularx}
\label{tab:HumanVsAgentTrajectory}
\end{table*}

\begin{definition}[\textbf{Repetitiveness Rate}] Quantifies the proportion of redundant actions within a task trajectory. An action is considered repetitive if it is identical to the action taken in the previous step.
\begin{equation*}
\textbf{Repetitiveness Rate} =
1-\frac{1}{\# \text{tasks}}\sum_{t \in \text{tasks}}\frac{\text{\# repetitive actions in task $t$}}{\text{\# actions in task $t$}}
\end{equation*}
\end{definition}    

In the raw form of the formula, without subtracting from 1, the fraction represents the proportion of repeated actions in the trajectory. A higher value of this fraction therefore indicates more repetition, which reflects worse performance. To ensure consistency across all metrics, where higher values indicate better performance, we invert the measure by subtracting this fraction from 1. This transformation maps a score of 1 to no repetitions and a score of 0.5 to trajectories that contain repetitions of all steps. In the example in Table~\ref{tab:HumanVsAgentTrajectory}, the agent repeated an action twice by clicking the Smartphones filter twice (A5, A6). The agent performed a total of six actions, resulting in a Repetitiveness Rate of $1-\frac{2}{6} \approx 0.66$.

\begin{definition}[\textbf{Step Success Rate}]
Calculates the proportion of human gold steps that are successfully fulfilled by the agent’s trajectory.
    \begin{equation*}
    \textbf{Step Success Rate} = 
    \frac{1}{\# \text{tasks}}\sum_{t \in \text{tasks}}\frac{\text{\# matched human gold steps in task $t$}}{\text{\# human gold steps in task $t$}}
    \end{equation*}
\end{definition}

In Table~\ref{tab:HumanVsAgentTrajectory}, the Step Success Rate is $\frac{3}{3} = 1.0$, indicating that the agent successfully completed all human-defined steps (A2, A3, A5).

\begin{definition}[\textbf{Partial Success Rate}] Measures how well the agent’s final output meets the specific requirements of a given task. When only a single element is required for a task, this metric is not applicable (in order of not extra-rewarding agents that fulfill a task with one output), hence we denote req\_tasks as the set of tasks with multiple requirements.
\begin{equation*}
    \textbf{Partial Success Rate} =
    \frac{1}{\# \text{req\_tasks}}\sum_{t \in \text{req\_tasks}}\frac{\text{\# completed requirements in task $t$}}{\text{\# requirements in task $t$}}
\end{equation*}
\end{definition}

This metric is useful for tasks with multiple correct elements. For example, in the task \textit{``Which US states border Connecticut?''}, the correct answer must include \textit{Rhode Island}, \textit{Massachusetts}, and \textit{New York}. If the agent produces only \textit{Massachusetts}, the Partial Success Rate is $\frac{1}{3}$. 
In the WebArena dataset, each task is annotated with a reference answer corresponding to a specific intent. Some tasks require the output to contain multiple values (e.g., an unordered list of text items) for the answer to be considered correct. We incorporate this information into our code implementation. 

\begin{definition}[\textbf{Element Accuracy Rate}] Compares the action the agent planned to take with the action it actually performed.     
\begin{equation*}
    \textbf{Element Accuracy Rate} =
    \frac{1}{\# \text{tasks}}\sum_{t \in \text{tasks}}\frac{\text{\# matching steps predicted and actual}}{\text{\# agent steps}}
\end{equation*}
\end{definition}

Occasionally, during the reasoning step, the agent explicitly expresses an intention to perform a specific task. For example, ``In summary, the next action I will perform is type [522] AMC Waterfront.'' However, the action that is ultimately executed may differ, or not even happen at all. To capture this gap, we propose the Element Accuracy Rate. In Table~\ref{tab:HumanVsAgentTrajectory}, the last step differs between the planned and actual actions (A6, AP6), yielding an Element Accuracy Rate of $\frac{5}{6}$.

\subsection{LLM-as-A-Judge in Our Work}
Our metrics rely on LLMs to measure the metric values. The LLM is being utilized mostly as a \textbf{semantic judge} that understands the action's written intent, even when the specific phrasing or format of actions differs. To implement this, we adapt the methodology proposed by \citet{liu2023g} for prompt-based evaluation. The prompt includes three elements: 
\begin{enumerate}
    \item A definition of the evaluation task to assess the semantic equivalence between the human's gold action and the agent's executed action.
    \item An evaluation criteria that establish the rules for scoring, e.g., the definition of semantic equivalence.
    \item Chain-of-Thought (CoT) instructions that provide the LLM with a sequential set of internal instructions, guiding it through the analysis process to ensure consistency \citep{wei2022chain}.
    \end{enumerate}
An example of the prompts that implement the semantic judge is provided in Table~\ref{tab:semantic_judge}.

\begin{table*}[h]
\centering
\caption{Example prompt structure for semantic equivalence.}\label{tab:semantic_judge}
\small
\begin{tabularx}{\textwidth}{|l|X|}
\hline
Prompt Component & Content and Purpose \\
\hline
\textbf{Task Definition} & You will be provided with two actions: 1. Human Recorded Action: [The human's action description] 2. Agent-Recorded Action: [The agent's action description, which may include metadata like IDs]. Your task is to determine  whether the Human-Recorded Action and the Agent-Recorded Action are semantically equivalent. \\
\hline
\textbf{Evaluation Criteria} & Semantic Equivalence means both actions express the same intent.
Your output must be a single numerical value:
1 if both actions are semantically identical, 0 if they are not.
Note:
1. Minor textual variations such as differences in casing (`Design' vs `design'), extra/missing spaces, or punctuation should be ignored as long as the core identity and meaning are preserved.
2. Any substantive mismatch in meaning or details (e.g., different dates) must result in a 0 \\
\hline
\textbf{Chain-of-Thought (CoT)} & Evaluation Steps: 1. Read the Human-Recorded Action and the Agent-Recorded Action carefully. 2. Rate the output as 1 (same) or 0 (different) according to the criteria above.
Example:  Human-Recorded Action: search for flights. Agent-Recorded Action: find flights. Output: 1\\
\hline
\end{tabularx}
\end{table*}

We use the semantic LLM judge framework in many of the defined evaluation metrics. For example, the Recovery Rate metric uses semantic similarity to compare the agent’s actions with the human reference actions. If the actions differ, we conclude that there is a deviation from the human trajectory, and this deviation is recorded. Then, we move to the next agent action and compare it with the corresponding human action, again computing a similarity score.
Another example is the Step Success Rate metric. To calculate the proportion of human gold steps that are successfully fulfilled by the agent’s trajectory, we first need to identify which human actions are semantically similar to the agent's actions. To achieve this, we provide an LLM with the two trajectories of the agent and the human gold, and ask it to return a list of matching agent steps. We use a prompting approach similar to the one applied in the Recovery Rate technique. The full prompt can be found in Appendix~\ref{app:promptStepSuccessRate}. The prompts for the Repetitiveness Rate and Partial Success Rate are provided in Appendices~\ref{app:promptRepetitivenessRate} and~\ref{app:promptPartialSuccessRate}, respectively.

\section{Experimental Setting}
\label{sec:method}

In the previous sections, we introduced a taxonomy for classifying web agents based on their planning mechanisms, presented our Full-Plan-in-Advance agent, and outlined a set of evaluation metrics for assessing agent behavior. In this section, we integrate these components into an experimental setup. 


\subsection{Dataset}
We run experiments on the full set of 812 WebArena tasks \citep{zhou2023webarena}.\footnote{\url{https://webarena.dev/}.}
The tasks span five domains: e-commerce (OneStopShop), social media discussions (Reddit), collaborative software development (GitLab), content management systems (CMS), and navigation systems (OpenStreetMap). Each task includes a natural language instruction and a corresponding acceptance criterion specifying what constitutes a correct outcome. For instance, in task 80, the task asks \textit{What is the duration required to first walk from Massachusetts Institute of Technology to Harvard University, and then drive to Boston Logan International Airport?} with the correct answer being \textit{63 minutes}.






\subsection{Web Agents}
As mentioned earlier, WebArena is not only a benchmark but also a functioning web agent of the Step-by-Step type (see Section~\ref{sec:webAgentTaxonomy}).
In our experiments, we compare WebArena’s Step-by-Step agent with our Full-Plan-in-Advance agent.
Both agents interpret natural language instructions and reason about web content using LLMs. We experiment with GPT-4o-mini with a temperature of 1.0 to encourage exploration in the search space. When the temperature is high, dividing the logits by $T$ makes all the values smaller/closer together, spreading out the probability distribution more evenly. This means the model is more likely to pick less probable tokens, increasing randomness, and in our case, incentivize exploration.
We used a top-p parameter of 0.95, which gives the model more freedom to discover shortcuts or less common action sequences that may lead to more effective plans. Both agents operate under identical environmental conditions to ensure a fair comparison.

As for the agent-level parameters, we set the maximum number of state transitions to 30. In addition, execution is stopped if the same action is repeated more than three times on the same observation or if the agent generates three consecutive invalid actions. We also applied several changes to the original WebArena prompt presented by \citet{Zhou2024}. The changes are provided in Appendix~\ref{app:promptWebarenaAgent}.

\begin{figure*}
\centering
\includegraphics[height=0.35\textheight]{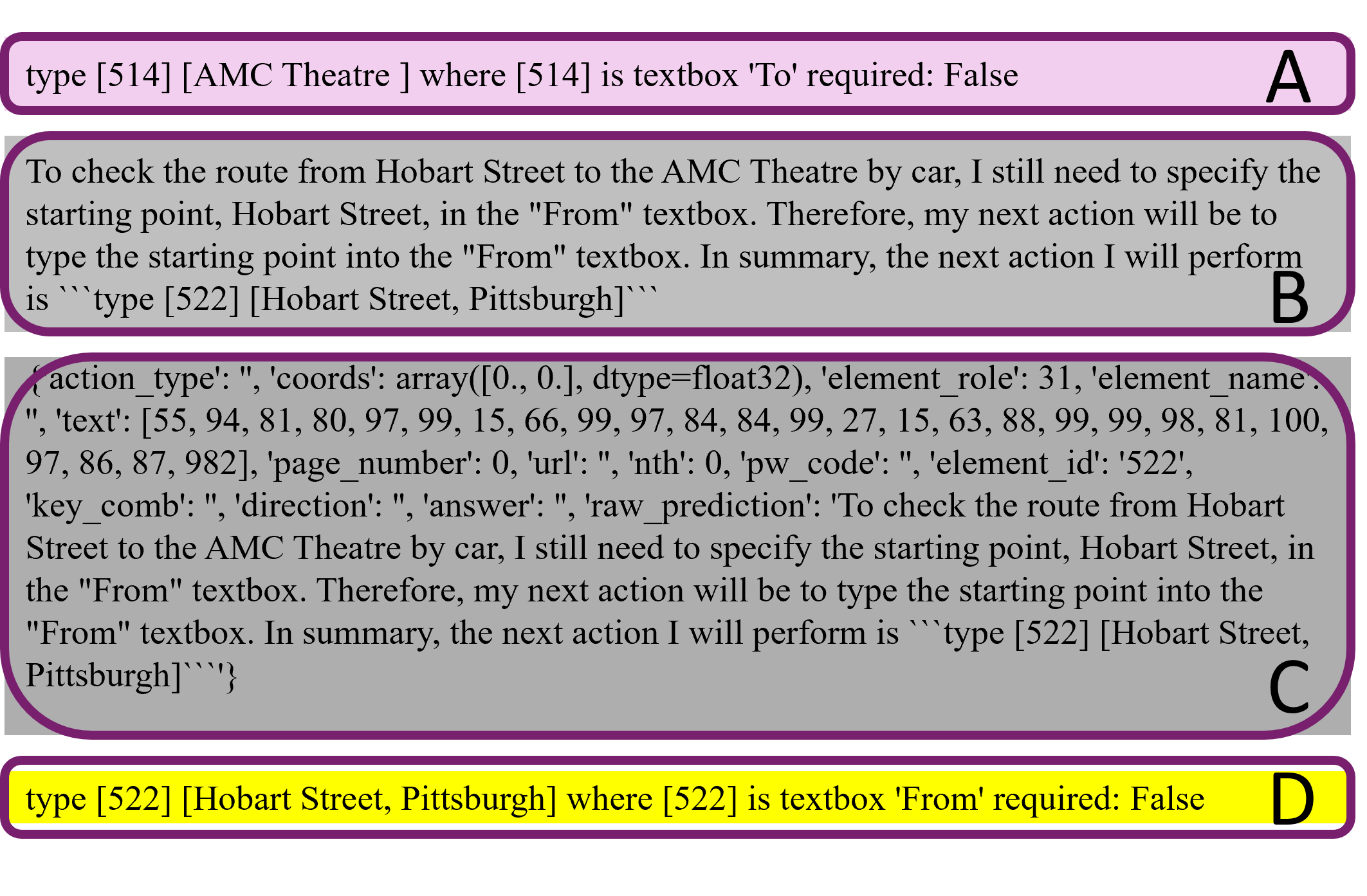}
\caption{An example step from task 40 illustrating the agent’s decision-making process. The \textbf{pink} section, labeled \textbf{A} represents the \textit{previous action}, the \textbf{top gray} section, labeled \textbf{B} details the agent’s \textit{reasoning process}, the \textbf{bottom gray} section, labeled \textbf{C}, contains \textit{meta data}, which we did not include in our analysis, and the \textbf{yellow} section, labeled \textbf{D} indicates the \textit{next action} to be performed.}
\label{fig:webarenaStepExample}
\end{figure*}

Once both agents completed the 812 tasks, the next stage of our methodology involves preparing their outputs for evaluation. The raw execution trajectories consist of HTML files, each representing a task divided into steps. Each step contains: (1) the previous action, (2) the agent’s reasoning for the next action, and (3) the next action suggestion. Using a Python script, we extracted these components from each HTML file and mapped them into a structured tabular database. Figure~\ref{fig:webarenaStepExample} provides an example of this HTML format and its corresponding components.
After constructing the tabular dataset, we applied the five evaluation metrics to every trajectory. Each trajectory receives a score for step success, recovery, element accuracy, repetitiveness, and partial success. The resulting metric scores are computed using a dedicated Python script\footnote{The code is available at \href{https://github.com/shahnovsky/WebArena-Human-Trajectory-Dataset}{Metrics Implementation}.} and subsequently analyzed using IBM SPSS 27 for Windows.

\paragraph{Annotation Process of Human Gold Trajectory Dataset}\label{sec:dataset}
We now turn to the human reference dataset. This dataset serves as one of the inputs for the Recovery Rate and the Step Success Rate metrics. The human golden trajectories are compared against the agents' behavior. For 794 out of 812 tasks of the Webarena dataset, the authors of this paper manually created human-gold trajectories by completing each of the tasks and documenting every step taken during the process of completing the tasks. As the web agents are required to generate the next step in a structured format for the execution step, we recorded both the type of action and the name of the element on which the action was performed. Overall, we successfully completed 97.7\% of the tasks. The remaining 18 tasks were not completed because we were uncertain about how to complete them.\footnote{Some of the tasks in the WebArena benchmark are unsolvable by design.} The complete dataset is available at \href{https://github.com/shahnovsky/WebArena-Human-Trajectory-Dataset}{Dataset Link}.

\section{Results and Discussion}
\label{sec:results}

In this section, we report the results of our experiments with the WebArena agent and our Full-Plan-in-Advance agent on the WebArena benchmark.
For this evaluation, both agents were tasked with performing all 812 WebArena tasks \citep{zhou2023webarena}. We then applied the five evaluation metrics introduced earlier to compare the behavior and performance of the two agents. The results for each metric are presented below.

\begin{table*}[h]
\centering
\caption{Success rates and relative change of the Full-Plan-in-Advance and Webarena agents per domain.}
\label{tab:SR_per_domain_per method}
\begin{tabular}{|c|c|c|c|c|}
    \hline
    \textbf{Website} & \textbf{Webarena Agent} & \textbf{Full-Plan-in-Advance Agent} & \textbf{delta} \\
    \hline
    CMS &7.69\% & 3.85\% & -3.84\%\\
    Gitlab & 12.33\% & 10.96\% & -1.37\% \\
    Map & 10.39\% & 12.99\% & -2.60\% \\
    Reddit &  4.00\% & 8.00\% & +4.00\%\\
    e-commerce &  5.00\% & 2.51\% & +4.00\%\\
    \hline
    All &  38.41\% & 36.29\% & -2.12\%\\
    \hline
\end{tabular}
\end{table*}

\begin{figure*}[h]
\centering
\includegraphics[width=\textwidth]{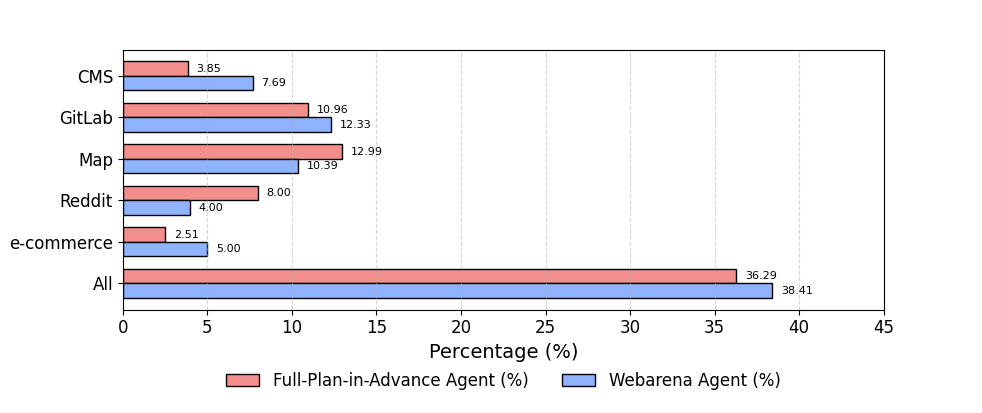}
\caption{Success rates of Step-by-Step agent and Full-Plan-in-Advance agent on the WebArena benchmark divided to success on each domain.}
\label{fig:success_results}
\end{figure*}

\subsection{Success Rate}
While our work introduces new metrics that capture the quality and structure of agent behavior as expressed by its planned and executed trajectories, we also report the traditionally and commonly reported success rate to allow a direct and fair comparison with prior work and existing benchmarks. Table~\ref{tab:SR_per_domain_per method} and Figure~\ref{fig:success_results} present the success rates of both agents across the five WebArena domains. The WebArena agent achieves an overall Success Rate of 38.41\%, while the Full-Plan-in-Advance agent obtains 36.29\%, reflecting a modest decrease in aggregate performance. At the domain level, the Full-Plan-in-Advance agent performs slightly worse on CMS, GitLab, and Map, with relative drops of –3.84\%, –1.37\%, and –2.60\%, respectively. In contrast, it shows improvements on Reddit and e-commerce, where success rates increase by +4\% in both cases. It is important to emphasize that the Full-Plan-in-Advance agent was not optimized for performance, nor was achieving state-of-the-art results a goal of this work. Rather, the agent is presented solely as a proof of concept and to illustrate the baseline performance of this class of agents.

These mixed results indicate that planning ahead may benefit certain domains, particularly those involving structured and predictable pages. Both E-commerce and Reddit domains share multiple common characteristics: e-commerce pages typically present product information, category listings, and search results in a rigidly structured layout. They feature predictable elements such as title, image, price, description, reviews, and a `Add to Cart' button. Likewise, subreddit post listings follow a consistent format, including title, author, subreddit name, upvote count, time posted, and a `Comment' button. This highly structured and uniform presentation makes it easier for the agent to generate a correct plan in advance, as the expected elements and their roles are relatively stable. When the agent’s plan aligns with this predictable structure, execution is more reliable.

\begin{figure*}[h]
\centering
\includegraphics[width=\textwidth]{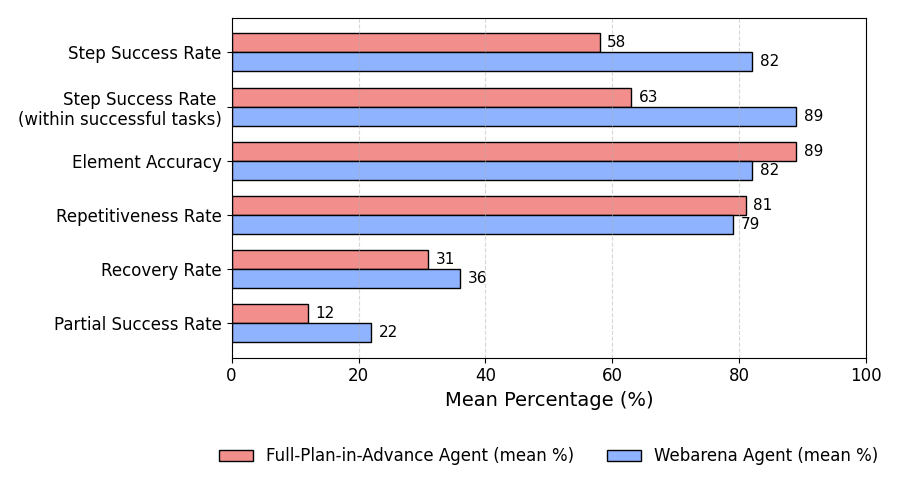}
\caption{Results of the proposed evaluation metrics of Step-by-Step agent and Full-Plan-in-Advance agent on the WebArena benchmark averaged over all domains.}
\label{fig:metrics_results}
\end{figure*}

Overall, success rate alone offers only a partial view, motivating the use of the additional trajectory level metrics introduced in this paper. Table~\ref{tab:AvarageScoreAllMetrics} and Figure~\ref{fig:metrics_results} present a summary of the new evaluation metric values, which capture different aspects of the agent's execution. 
It provides a summary of the scores across all metrics for both agents, and the following sections present a detailed analysis of each metric in turn. Note that although the table reports metric scores on a 0–1 scale, we usually describe the results in percentages for readability (e.g., 0.82 is reported as 82\%).

\begin{table*}[t]
\centering
\caption{Average score for all metrics.}
\label{tab:AvarageScoreAllMetrics}
\begin{tabularx}{\textwidth}{|l|X|X|}
    \hline
    \textbf{Evaluatin Metric} & \textbf{Webarena Agent} & \textbf{Full-Plan-in-Advance Agent}   \\
    \hline
    Step Success Rate & $0.82 \pm 0.14$ & $0.58 \pm 0.29$ \\
    Step Success Rate (within successful tasks) & $0.89\pm 0.06 $ & $0.63 \pm 0.10 $ \\
    Element Accuracy & $0.82 \pm 0.12$ & $0.89 \pm 0.03$ \\
    Repetitiveness Rate & $0.79 \pm 0.14$ & $0.81 \pm .13$ \\
    Recovery Rate &    $0.36 \pm 0.19$ & $0.31 \pm 0.12$\\
    Partial Success Rate & $0.22 \pm 0.39$ & $0.12 \pm 0.27$ \\
    \hline
\end{tabularx}
\end{table*}

\subsection{Step Success Rate}
From the new metrics, we first examine the step success rate, which measures how closely each agent’s sequence of actions aligns with the human gold-reference trajectory.  
As shown in Table~\ref{tab:AvarageScoreAllMetrics}, the average step success rate for the Full-Plan-in-Advance agent is 58\%, compared to 82\% for the WebArena agent. This substantial gap shows that our agent frequently diverges from the human gold trajectory. This can occur because the LLM-generated plan is not structured in a way that reflects human reasoning. Table~\ref{tab:82HumanVSgenralPlan} illustrates this clearly: in a task that requires changing the travel mode to \textit{foot} before proceeding, the agent’s plan does not include this essential step, likely because it assumes that all standard travel modes are already available simultaneously. In contrast, a human observes the user interface directly, immediately notices that the default transportation mode is not set to \textit{foot}, and therefore chooses to click the relevant button to update it before continuing. In addition, the agent introduces unnecessary actions, such as instructing itself to extract information from the screen, a process that is trivial for humans and therefore never explicitly planned in the human trajectory. Overall, the Full-Plan-in-Advance agent’s lower step success rate may be attributed to its inability to predict in advance what it will encounter and either trying to prepare for the worst case scenario or assuming the best case scenario. This pattern was also occurring when only considering tasks that the agent successfully completed, as shown in the second row of Table~\ref{tab:AvarageScoreAllMetrics}.

\begin{table*}[h]
\centering
\caption{Human trajectory for Task 82 alongside the general plan produced by the Full-Plan-in-Advance agent. Differences between the two appear in bold text.}
\label{tab:82HumanVSgenralPlan}
\begin{tabularx}{\textwidth}{|c|X|X|}
    \hline
    \multicolumn{3}{|l|}{\textbf{Task 82:} What is the duration required to first walk from Massachusetts Institute of Technology to} \\ 
    \multicolumn{3}{|l|}{Harvard University, and then drive to Boston Logan International Airport?} \\
    \hline
    \textbf{\#} & \textbf{Human Trajectory} & \textbf{General Plan by Full-Plan-in-Advance agent}  \\
    \hline
    1 & Click on the 'Find directions between two points' & Click on the link 'Find directions between two points'  \\
    \hline
    2 & Type 'Massachusetts Institute of Technology' into the 'From' textbox & Type "Massachusetts Institute of Technology" in the "From" textbox \\ 
    \hline
    3 & Type 'Harvard University' into the 'To' textbox & Type "Harvard University" in the "To" textbox \\
    \hline
    4 & \textbf{Select the 'Foot (OSRM)' option} & \\
    \hline
    5 & Click the 'Go' button & Click on the button 'Go' \\
    \hline
    6 & Click the 'Reverse Directions' button & \textbf{Note down the walking duration information} \\
    \hline
    7 &  & Type "Harvard University" in the "From" textbox \\
    \hline
    8 & Type 'Boston Logan International Airport' into the 'To' textbox & Type "Boston Logan International Airport" in the "To" textbox  \\
    \hline
    9 & Click the 'Go' button &  Click on the button 'Go' again \\
    \hline
    10 & & \textbf{Note down the driving duration information} \\
    \hline
\end{tabularx}
\end{table*}


While computing the Step Success Rate for each task, the evaluation code also exposes additional information that is not part of the metric itself; Specifically, the lengths of both the human trajectory and the agent’s trajectory. This auxiliary information provides insight into the efficiency of each agent, as longer trajectories may indicate unnecessary or redundant steps. Table~\ref{tab:StepsNumber} summarizes the average number of steps observed in trajectories produced by humans and agents. Human trajectories required an average of $7.92 \pm 5.18$ steps, which is lower than the average number of steps of both the Full-Plan-in-Advance agent and the WebArena agent, $20.21 \pm 10.16$, and $15.02 \pm 8.93$, respectively. This increase suggests both agents often performs additional actions beyond what is necessary to complete the task.


\begin{table}[h]
\centering
\caption{Average number of steps (mean ± SD) observed across human golden trajectories, WebArena agent, and Full-Plan-in-Advance agent trajectories.}
\label{tab:StepsNumber}
\begin{tabular}{|l|c|}
    \hline
    \textbf{Agent} & \textbf{Number of Steps} \\
    \hline
    Human & $7.92 \pm 5.18$ \\
    \hline
    WebArena Agent & $15.02 \pm 8.93$ \\
    \hline
    Full-Plan-in-Advance Agent & $20.21 \pm 10.16$\\
    \hline
\end{tabular}

\end{table}

\subsection{Element Accuracy Rate}
\begin{figure*}[h]
\centering
\includegraphics[height=0.35\textheight]{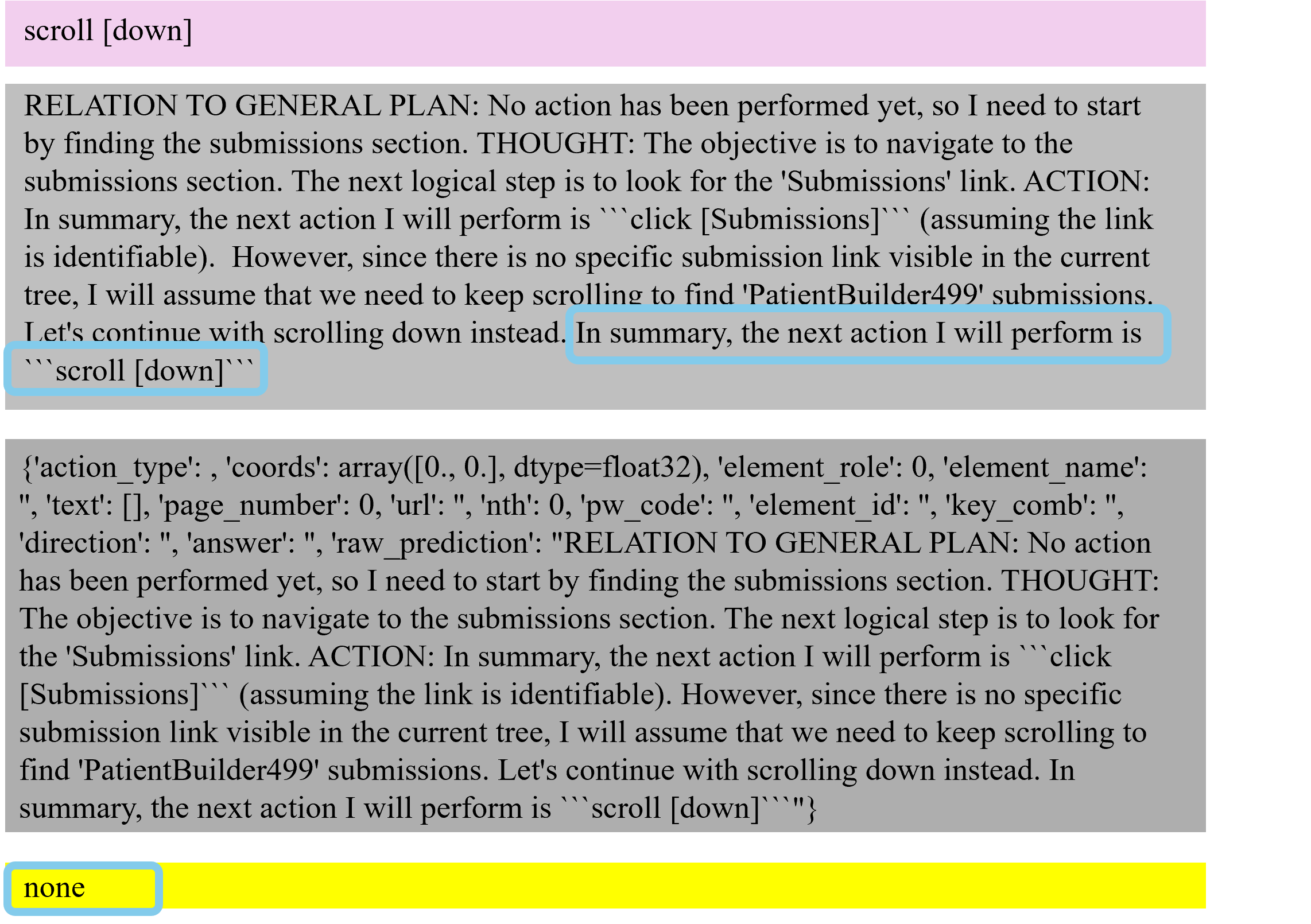}
\caption{A step performed by the Full-Plan-in-Advance agent for task number 727, where the reasoning phase suggests scrolling in the correct format of the instruction, but the actual action taken in `None'. \textbf{Task 727}: DisLike all submissions created by PatientBuilder499 in subreddit videos.}
\label{fig:727plan_2}
\end{figure*}

The element accuracy rates, which compares the step an agent planned to take with the action it actually performed, for the Full-Plan-in-Advance agent and the WebArena agent, are 89.89\% and 82.45\%, respectively. These results indicate that sometimes the agent’s difficulty lies not in identifying the correct element, but in the execution phase. Although the agents typically select the right target element, it sometimes fails to successfully perform the action on the first attempt. Carefully observing the trajectories, in some cases, the action eventually succeeds after one or two retries; in others, the agent repeatedly attempts the same action until it reaches the maximum number of allowed identical attempts (a configured parameter). Once this limit is reached, the agent terminates and the task is marked as failed, even though the correct element was identified. For instance, as illustrated in Figure~\ref{fig:727plan_2}, even though the agent selects a valid action, it is unable to execute it for reasons that are not observable.

\begin{figure*}[h]
\centering
\includegraphics[height=0.35\textheight]{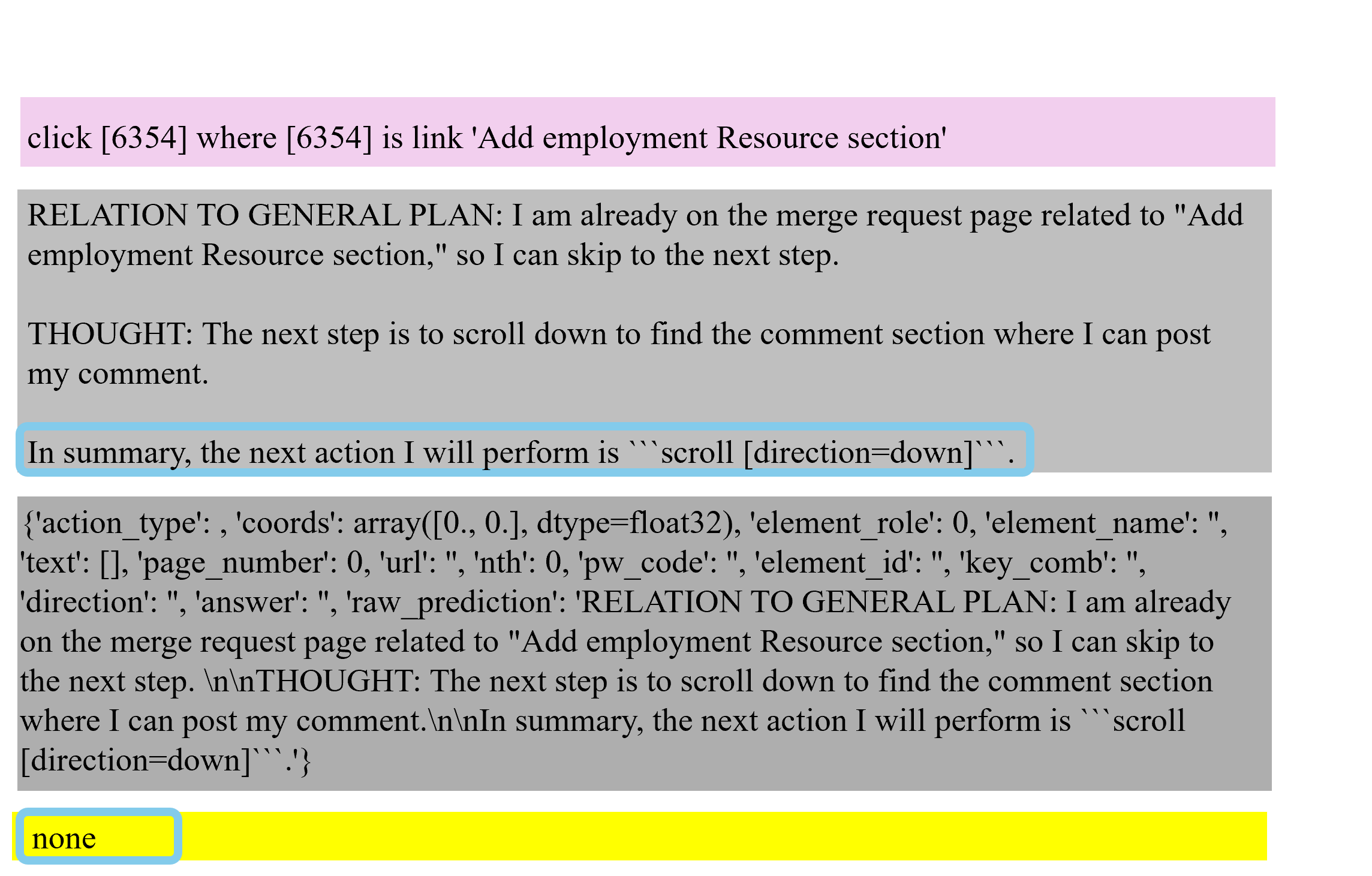}
\caption{A step performed by the Full-Plan-in-Advance agent for task 392, where the reasoning phase suggests scrolling but does not follow the required action format, causing the actual executed action to be `None'. \textbf{Task 392}: Post ``Good idea'' for the merge request related to color ulitity in a11yproject.com project.}
\label{fig:727plan}
\end{figure*}

Another source of mismatch between the proposed action and the executed one is the format suggested by the reasoning mechanism. Although the system prompts instruct both agents to strictly follow the required action format, the LLM does not always comply. As shown in Figure~\ref{fig:727plan}, the agent attempts to perform a scroll-down action, but expresses it in an incorrect format. For scroll actions, the valid forms are \texttt{scroll [down]} or \texttt{scroll [up]}, rather than \texttt{scroll [direction=down]} or \texttt{scroll [direction=up]}.

\subsection{Repetitiveness Rate} 
We next evaluate how many times an agent tries to perform the same action repeatedly. The repetitiveness rate for the Webarena agent is 79\%, which means that on average 21\% of the actions were constantly repeated. Full-Plan-in-Advance performs slightly better, as only 19\% of its actions are repetitive. These repetitions are not only inefficient, but they directly affect task completion as execution is halted if the same action is repeated more than three times on the same observation, or if three consecutive invalid actions are generated. These are safety mechanisms of both agents. 

Interestingly, although the Full-Plan-in-Advance agent produces more steps overall (see Table~\ref{tab:StepsNumber}), it generates fewer repeated actions. This suggests that its failures often arise from plans that drift away from the correct sequence of golden human steps (for example, adding irrelevant actions), rather than from repeatedly attempting the same action. 
In contrast, the WebArena agent repeats the same action more frequently without making any progress.

\subsection{Recovery Rate}
We further analyze the recovery rate to evaluate each agent’s ability to recover after deviating from the golden human trajectory. This metric captures how many recoveries were detected among the deviations, compared to the full human golden trajectory.
The WebArena agent recovers from deviations in roughly 36\% of deviation incidents on average. The relatively high standard deviation (0.19) indicates that its recovery ability varies substantially across tasks. The Full-Plan-in-Advance agent recovers slightly less often, 31\% on average. 

What stands out in Tables~\ref{tab:SR_per_domain_per method} and~\ref{tab:AvarageScoreAllMetrics} is the noticeable alignment between the Overall Success Rate and the Recovery Rate. The Full-Plan-in-Advance agent completes 36.29\% of the tasks successfully, and its recovery rate is 31\%, which is very close to its overall performance. This suggests that when the agent manages to recover from a deviation, it is often able to complete the task.
Similarly, the WebArena agent completes 38.41\% of the tasks successfully and its recovery rate of 36\% is also close to its overall success rate. This indicates that, for both agents, recovery tends to coincide with task completion.



\subsection{Partial Success Rate}
Finally, we report the partial success rate, which evaluates how well each agent’s final answer satisfies tasks with multiple required elements. The WebArena agent attains an average score of $0.22\% \pm 0.39$, whereas the Full-Plan-in-Advance agent reaches $0.12\% \pm 0.27$. These values are relatively low for both agents, indicating that in many cases, even when the task is only partially solved, the responses cover very few of the required elements. The large standard deviations further suggest substantial variability across tasks. 
Both agents occasionally identify many of the required elements, but more often their outputs include only a small portion of what the task demands, and in some cases none of the required elements appear at all.

To further understand how each agent behaves on tasks that require multiple correct elements in the final response, we analyze the types of outcomes produced. While the partial success rate quantifies how many required elements appear in the final output, it does not reveal whether the agent reached the answer generation stage, stopped prematurely, or failed to produce any answer at all. We logged this information separately, not as part of the metric itself, but because it was already provided by the LLM during execution and therefore available to record. Table~\ref{tab:partialSuccesAnwers} categorizes each outcome into three groups: tasks for which no answer was produced (N/A), tasks that ended in an early stop, and tasks for which the agent generated a final incomplete answer. 

The category Early Stop refers to cases in which the agent terminates execution before producing a final answer. This can occur for three distinct reasons:
First, the agent may repeat the same action three consecutive times, triggering the mechanism that halts execution; this threshold is a user-controlled parameter. Second, the agent may produce three consecutive unparsable or invalid actions, after which execution is stopped because the agent is no longer generating commands that it can interpret and execute. Finally, an early stop is recorded when the agent reaches the maximum allowed number of steps (30 by default, also configurable by the user), indicating that it could not complete the task within the predefined limit. These conditions help identify situations where the agent is unable to make further progress and serve both as loop-prevention mechanisms and cost-control measures.

The N/A category represents tasks for which the agent did not produce any final answer. Importantly, this outcome is considered legitimate because the WebArena benchmark includes 161 tasks that are inherently unachievable \citep{trymeka2025webarena}. In our annotations, we were unable to complete less tasks, only 18 of them. Therefore, interpreting N/A counts must take into account that some tasks are unsolvable by design.

Table~\ref{tab:partialSuccesAnwers} shows that the WebArena agent produced answers for 55 tasks which have more than one output, whereas the Full-Plan-in-Advance agent produced answers for 48, indicating that the planning-based approach is less likely to reach the answer-generation stage. Both agents show a comparable number of N/A results (17 and 20, respectively), which is expected given that some tasks are unachievable regardless of strategy. The Full-Plan-in-Advance agent exhibits a slightly higher number of early stops, 28 vs. 24 for the WebArena agent, consistent with earlier observations that its rigid plans can lead to execution failure.

Considering the rate of legitimate answers, which consists of the N/A and Produced Answer categories, we can better understand the underlying reasons for agent failure. A substantial portion of failures does not stem from the impossibility of the task itself or from following a wrong trajectory, but rather from the agent’s own behavior during execution. In particular, the remaining failures are caused by early stops due to repeated actions, invalid commands, or exceeding the step limit. This suggests that the agents often fail because they are unable to progress reliably through the task execution process.

\begin{table*}[h]
\centering
\caption{Distribution of final-answer types for tasks requiring multiple correct elements, comparing how often each agent produced no answer (N/A), stopped early, or generated a final answer.}
\label{tab:partialSuccesAnwers}
\begin{tabular}{|l|l|l|}
    \hline
    \textbf{Answer Category} & \textbf{WebArena Agent} & \textbf{Full-Plan-in-Advance Agent} \\
    \hline
    N/A & 17 & 20 \\
    Produced Answer & 55 & 48 \\
    \hline
    \textbf{Percentage of Legitimate Answers} & \textbf{75.00\%} & \textbf{70.83\%} \\
    \hline
    Early Stop & 24 & 28 \\
    \hline
    Sum & 96 & 96 \\
    \hline
\end{tabular}
\end{table*}

\section{Discussion} 
Across all evaluation metrics, the WebArena agent and the Full-Plan-in-Advance agent display different behavioral patters, each with distinct strengths and limitations.
The WebArena agent, categorized as a Step-by-Step agent, demonstrated stronger alignment with the human reference trajectory, reflected in its higher step success rate and higher recovery rate. These results indicate that it is more stable during execution and better able to realign after deviations. The agent also exhibited a shorter average trajectory length.
The Full-Plan-in-Advance agent, which falls under the same category name, demonstrated advantages in more technical aspects. It achieved higher element accuracy rate and produced slightly fewer repeated actions. However, these strengths do not translate into stronger task-level performance. The planning agent consistently underperforms across metrics that depend on sustained sequential correctness: step success rate, recovery rate, partial success rate, and ultimately overall success rate.

A closer look at the agents behavior during execution helps explain this pattern. The planning agent frequently omits essential steps or introduces unnecessary ones, causing its trajectories to diverge from the human path early. Once deviated, it struggles to recover and often terminates due to repeated actions, invalid commands, or exceeding the step limit. The analysis of final-answer outcomes reinforces this---the Full-Plan-in-Advance agent reaches the answer generation phase less often and encounters early stops more frequently.

To develop more effective web agents, it is necessary to identify which classes of web-based tasks benefit from different planning strategies. While some dynamic web environments demand constant reaction to newly observed states, others allow the agent to rely on structured reasoning and predefined navigational sequences.

\paragraph{When to Use Step-by-Step Planning} 
The Step-by-Step planning paradigm, which employs a breadth-oriented traversal of the immediate action space, is most suitable for web environments that are dynamic, data-dependent, or partially observable. Such tasks require the agent to repeatedly assess the current state of the Document Object Model (DOM) before deciding on the next action.
This approach is particularly relevant for web-based cloud management consoles and DevOps dashboards (e.g., GitLab web interfaces). In these settings, state transitions are often slow and non-deterministic; for example, a web-based deployment pipeline may require several minutes to initialize, and its final status cannot be assumed in advance. As a result, the agent must observe the system's health on the web page before determining whether it is safe to proceed with the next click or input. 
In these scenarios, the agent's ability to recover from failure becomes a central indicator of performance. If a web element fails to load, a high-quality agent should realign its behavior by refreshing or navigating back. A process-aware evaluation framework would penalize agents that repeatedly issue the same failing command (high Repetitiveness Rate) and reward those that successfully adapt their trajectory toward a safe execution path.

Another critical application arises in web-based Electronic Health Record (EHR) portals or dynamic social media forums (e.g., Reddit). In these workflows, the presence of specific web elements depends heavily on user-generated content or external database confirmations. The agent cannot predict the exact layout of the next page, making step-by-step verification safety-critical. Measuring the gap between the agent's reasoning phase and its execution phase provides a necessary safety layer for these unpredictable web interfaces.

\paragraph{When to Use Full-Plan-in-Advance} 
The Full-Plan-in-Advance paradigm, characterized by a depth-oriented traversal of a pre-calculated trajectory, is most effective in web environments that are highly structured, uniform, and governed by rigid business logic. In such settings, the agent can exploit the predictability of the web interface to generate a comprehensive execution roadmap prior to taking any action.
Enterprise web systems (e.g., SAP or Oracle NetSuite web portals) and standard e-commerce platforms exemplify this category. E-commerce pages typically present product information, category listings, and checkout pipelines in a rigidly structured layout featuring predictable elements. A typical automated workflow requires the agent to log into the portal, navigate a static menu hierarchy, match line items, and trigger payment approval. 
These environments strongly favor pre-planned execution. While a Step-by-Step agent may become distracted by a large number of available menu options or promotional pop-ups, a Full-Plan-in-Advance agent benefits from following a predefined trajectory that aligns with corporate audit requirements.

A similar need for structured planning appears in web-based Content Management Systems (CMS). Consider the task of publishing a scheduled post with specific formatting and tags. The hierarchy of operations (e.g., navigating to the dashboard, opening the editor, inserting media, setting the publication date, and clicking publish) follows strict ordering constraints. Pre-planning enables the agent to validate the entire logical sequence before interacting with the live web forms, thereby reducing the risk of incomplete or improperly formatted publications.

\section{Conclusion}

In this paper, we address multiple challenges that refer to the development and evaluation of autonomous web agents. Modern LLM-based agents have demonstrated remarkable capabilities on web tasks, yet their decision making processes often remain opaque, and existing evaluation procedures focus almost exclusively on task-completion success rates, overlooking the quality of the execution process itself, hence hindering the understanding of the way these agents operate. 

To better understand how web agents work and how to develop improved agents, we introduce a framework that connects LLM-based agent designs with classical planning methodologies. By formally mapping web agents to established search paradigms: Step-by-Step to BFS, Tree Search to Best-First Tree-Search, and Full-Plan-in-Advance to DFS, we provide a foundation for comparing agent architectures and a path for developing new types of agents. 
Importantly, the principles of this classification extend beyond web browsing. Since the taxonomy is derived from the state-space search, a concept applicable across all domains, it can be generalized to areas where agents must plan and act, including robotics, GUI automation, multimodal systems, and real-world task execution.

Building on this, we further propose a suite of five evaluation metrics designed to capture the coherence and efficiency of an agent’s execution trajectory: Step Success Rate, Element Accuracy, Repetitiveness Rate, Recovery Rate, and Partial Success Rate. Unlike traditional binary success metrics, our metrics assess the intermediate decision making process. These metrics, while designed and validated on WebArena, are not domain specific; they can be applied to any setting where agents produce multi-step trajectories.
To support the process of the evaluation, we construct a human gold-annotated trajectory dataset containing 794 expert-annotated execution traces across the WebArena benchmark. 

We conclude by using our framework to compare two agents with different planning methods: one that plans step-by-step and another that implements a full-plan-in-advance paradigm, the latter is our own novel implementation since such an agent was not implemented prior to this work. Our results show that the planning method of the agent significantly affects performance. The Step-by-Step agent (WebArena) follows the human path better, achieving a higher step success rate and recovery rate. In contrast, the Full-Plan-in-Advance agent is better on the technical side, showing higher element accuracy and repetitivness rate.

\bibliographystyle{ACM-Reference-Format}
\bibliography{sample-base}

@String{Computer = "{IEEE} Computer" }

@String{Springer = "Springer-Verlag" }

@article{liu2023llm+,
  title={Llm+ p: Empowering large language models with optimal planning proficiency},
  author={Liu, Bo and Jiang, Yuqian and Zhang, Xiaohan and Liu, Qiang and Zhang, Shiqi and Biswas, Joydeep and Stone, Peter},
  journal={arXiv preprint arXiv:2304.11477},
  year={2023}
}

@article{yao2023tree,
  title={Tree of thoughts: Deliberate problem solving with large language models},
  author={Yao, Shunyu and Yu, Dian and Zhao, Jeffrey and Shafran, Izhak and Griffiths, Tom and Cao, Yuan and Narasimhan, Karthik},
  journal={Advances in neural information processing systems},
  volume={36},
  pages={11809--11822},
  year={2023}
}

@inproceedings{xin2025bfs,
    title = "{BFS}-Prover: Scalable Best-First Tree Search for {LLM}-based Automatic Theorem Proving",
    author = "Xin, Ran  and
      Xi, Chenguang  and
      Yang, Jie  and
      Chen, Feng  and
      Wu, Hang  and
      Xiao, Xia  and
      Sun, Yifan  and
      Zheng, Shen  and
      Ding, Ming",
    booktitle = "Proceedings of the 63rd Annual Meeting of the Association for Computational Linguistics (Volume 1: Long Papers)",
    month = jul,
    year = "2025",
    address = "Vienna, Austria",
    publisher = "Association for Computational Linguistics",
    url = "https://aclanthology.org/2025.acl-long.1565/",
    doi = "10.18653/v1/2025.acl-long.1565",
    pages = "32588--32599",
    ISBN = "979-8-89176-251-0"
}

@inproceedings{he2024webvoyager,
    title = "{W}eb{V}oyager: Building an End-to-End Web Agent with Large Multimodal Models",
    author = "He, Hongliang  and
      Yao, Wenlin  and
      Ma, Kaixin  and
      Yu, Wenhao  and
      Dai, Yong  and
      Zhang, Hongming  and
      Lan, Zhenzhong  and
      Yu, Dong",
    booktitle = "Proceedings of the 62nd Annual Meeting of the Association for Computational Linguistics (Volume 1: Long Papers)",
    month = aug,
    year = "2024",
    address = "Bangkok, Thailand",
    publisher = "Association for Computational Linguistics",
    url = "https://aclanthology.org/2024.acl-long.371/",
    doi = "10.18653/v1/2024.acl-long.371",
    pages = "6864--6890"
}

@article{pan2024autonomous,
  title={Autonomous evaluation and refinement of digital agents},
  author={Pan, Jiayi and Zhang, Yichi and Tomlin, Nicholas and Zhou, Yifei and Levine, Sergey and Suhr, Alane},
  journal={arXiv preprint arXiv:2404.06474},
  year={2024}
}

@article{wei2022chain,
  title={Chain-of-thought prompting elicits reasoning in large language models},
  author={Wei, Jason and Wang, Xuezhi and Schuurmans, Dale and Bosma, Maarten and Xia, Fei and Chi, Ed and Le, Quoc V and Zhou, Denny and others},
  journal={Advances in neural information processing systems},
  volume={35},
  pages={24824--24837},
  year={2022}
}

@inproceedings{liu2023g,
    title = "{G}-Eval: {NLG} Evaluation using Gpt-4 with Better Human Alignment",
    author = "Liu, Yang  and
      Iter, Dan  and
      Xu, Yichong  and
      Wang, Shuohang  and
      Xu, Ruochen  and
      Zhu, Chenguang",
    booktitle = "Proceedings of the 2023 Conference on Empirical Methods in Natural Language Processing",
    month = dec,
    year = "2023",
    address = "Singapore",
    publisher = "Association for Computational Linguistics",
    url = "https://aclanthology.org/2023.emnlp-main.153/",
    doi = "10.18653/v1/2023.emnlp-main.153",
    pages = "2511--2522"
}

@article{xue2025illusion,
  title={An illusion of progress? assessing the current state of web agents},
  author={Xue, Tianci and Qi, Weijian and Shi, Tianneng and Song, Chan Hee and Gou, Boyu and Song, Dawn and Sun, Huan and Su, Yu},
  journal={arXiv preprint arXiv:2504.01382},
  year={2025}
}

@inproceedings{huang2023chatgpt,
  title={Is chatgpt better than human annotators? potential and limitations of chatgpt in explaining implicit hate speech},
  author={Huang, Fan and Kwak, Haewoon and An, Jisun},
  booktitle={Companion proceedings of the ACM web conference 2023},
  pages={294--297},
  year={2023}
}

@article{gilardi2023chatgpt,
  title={ChatGPT outperforms crowd workers for text-annotation tasks},
  author={Gilardi, Fabrizio and Alizadeh, Meysam and Kubli, Ma{\"e}l},
  journal={Proceedings of the National Academy of Sciences},
  volume={120},
  number={30},
  pages={e2305016120},
  year={2023},
  publisher={National Academy of Sciences}
}

@inproceedings{Zhou2024,
  title = {WebArena: A Realistic Web Environment for Building Autonomous Agents},
  author = {Zhou, Shuyan and Xu, Frank F. and Zhu, Hao and Bisk, Yonatan and Neubig, Graham \emph{et al.}},
  booktitle = {International Conference on Learning Representations (ICLR)},
  year = {2024}
}

@misc{trymeka2025webarena,
  author = {trymeka},
  title = {{WebArena Evals for Meka v1}},
  howpublished = {\url{https://github.com/trymeka/webarena_evals}},
  year = {2025},
  note = {GitHub repository; accessed on 2025-11-15}
}

@inproceedings{Li2023,
  title = {A Zero-Shot Language Agent for Computer Control with Structured Reflection},
  author = {Li, Tao and Li, Gang and Deng, Zhiwei and Wang, Bryan and Li, Yang},
  booktitle = {Proceedings of the 2023 Conference on Empirical Methods in Natural Language Processing (EMNLP)},
  year = {2023}
}

@inproceedings{Song2024,
  title = {Trial and Error: Exploration-Based Trajectory Optimization for LLM Agents},
  author = {Song, Yifan and Yin, Da and Yue, Xiang and Huang, Jie and Li, Sujian and Lin, Bill Yuchen},
  booktitle = {Proceedings of the 62nd Annual Meeting of the ACL (Volume 1: Long Papers)},
  year = {2024}
}

@article{ferrag2025llm,
  title={From llm reasoning to autonomous ai agents: A comprehensive review},
  author={Ferrag, Mohamed Amine and Tihanyi, Norbert and Debbah, Merouane},
  journal={arXiv preprint arXiv:2504.19678},
  year={2025}
}

@article{zhou2023webarena,
  title={Webarena: A realistic web environment for building autonomous agents},
  author={Zhou, Shuyan and Xu, Frank F and Zhu, Hao and Zhou, Xuhui and Lo, Robert and Sridhar, Abishek and Cheng, Xianyi and Ou, Tianyue and Bisk, Yonatan and Fried, Daniel and others},
  journal={arXiv preprint arXiv:2307.13854},
  year={2023}
}

@inproceedings{gatto2023text,
    title = "Text Encoders Lack Knowledge: Leveraging Generative {LLM}s for Domain-Specific Semantic Textual Similarity",
    author = "Gatto, Joseph  and
      Sharif, Omar  and
      Seegmiller, Parker  and
      Bohlman, Philip  and
      Preum, Sarah M.",
    booktitle = "Proceedings of the Third Workshop on Natural Language Generation, Evaluation, and Metrics (GEM)",
    month = dec,
    year = "2023",
    address = "Singapore",
    publisher = "Association for Computational Linguistics",
    url = "https://aclanthology.org/2023.gem-1.23/",
    pages = "277--288",
}

@article{koh2024tree,
  title={Tree search for language model agents},
  author={Koh, Jing Yu and McAleer, Stephen and Fried, Daniel and Salakhutdinov, Ruslan},
  journal={arXiv preprint arXiv:2407.01476},
  year={2024}
}

@article{han2024llm,
  title={LLM multi-agent systems: Challenges and open problems},
  author={Han, Shanshan and Zhang, Qifan and Yao, Yuhang and Jin, Weizhao and Xu, Zhaozhuo},
  journal={arXiv preprint arXiv:2402.03578},
  year={2024}
}

@article{dongre2025drift,
  title={Drift No More? Context Equilibria in Multi-Turn LLM Interactions},
  author={Dongre, Vardhan and Rossi, Ryan A and Lai, Viet Dac and Yoon, David Seunghyun and Hakkani-T{\"u}r, Dilek and Bui, Trung},
  journal={arXiv preprint arXiv:2510.07777},
  year={2025}
}

@article{deng2023mind2web,
  title={Mind2web: Towards a generalist agent for the web},
  author={Deng, Xiang and Gu, Yu and Zheng, Boyuan and Chen, Shijie and Stevens, Sam and Wang, Boshi and Sun, Huan and Su, Yu},
  journal={Advances in Neural Information Processing Systems},
  volume={36},
  pages={28091--28114},
  year={2023}
}

@book{ghallab2004automated,
  title={Automated Planning: theory and practice},
  author={Ghallab, Malik and Nau, Dana and Traverso, Paolo},
  year={2004},
  publisher={Elsevier}
}

@book{lavalle2006planning,
  title={Planning algorithms},
  author={LaValle, Steven M},
  year={2006},
  publisher={Cambridge university press}
}

@article{blum1997fast,
  title={Fast planning through planning graph analysis},
  author={Blum, Avrim L and Furst, Merrick L},
  journal={Artificial intelligence},
  volume={90},
  number={1-2},
  pages={281--300},
  year={1997},
  publisher={Elsevier}
}

@book{haslum2019introduction,
  title={An introduction to the planning domain definition language},
  author={Haslum, Patrik and Lipovetzky, Nir and Magazzeni, Daniele and Muise, Christian and Brachman, Ronald and Rossi, Francesca and Stone, Peter},
  volume={13},
  year={2019},
  publisher={Springer}
}

@article{pryor1996planning,
  title={Planning for contingencies: A decision-based approach},
  author={Pryor, Louise and Collins, Gregg},
  journal={Journal of Artificial Intelligence Research},
  volume={4},
  pages={287--339},
  year={1996}
}

@inproceedings{zheng2024gpt,
author = {Zheng, Boyuan and Gou, Boyu and Kil, Jihyung and Sun, Huan and Su, Yu},
title = {GPT-4V(ision) is a generalist web agent, if grounded},
year = {2024},
publisher = {JMLR.org},
booktitle = {Proceedings of the 41st International Conference on Machine Learning},
articleno = {2538},
numpages = {37},
location = {Vienna, Austria},
series = {ICML'24}
}

@article{nogueira2016end,
  title={End-to-end goal-driven web navigation},
  author={Nogueira, Rodrigo and Cho, Kyunghyun},
  journal={Advances in neural information processing systems},
  volume={29},
  year={2016}}

@phdthesis{sirin2004automated,
  title={Automated composition of web services using AI planning techniques},
  author={Sirin, Evren},
  year={2004},
  school={University of Maryland College Park, MD, USA}
}

@article{el2012distributed,
  title={A distributed multi-agent planning approach for automated web services composition},
  author={El Falou, Mohamad and Bouzid, Maroua and Mouaddib, Abdel-Illah and Vidal, Thierry},
  journal={Web Intelligence and Agent Systems},
  volume={10},
  number={4},
  pages={423--445},
  year={2012},
  publisher={SAGE Publications Sage UK: London, England}
}

@article{sodhi2023step,
  title={Step: Stacked llm policies for web actions},
  author={Sodhi, Paloma and Branavan, SRK and Artzi, Yoav and McDonald, Ryan},
  journal={arXiv preprint arXiv:2310.03720},
  year={2023}
}

@inproceedings{zhang2025webpilot,
  title={Webpilot: A versatile and autonomous multi-agent system for web task execution with strategic exploration},
  author={Zhang, Yao and Ma, Zijian and Ma, Yunpu and Han, Zhen and Wu, Yu and Tresp, Volker},
  booktitle={Proceedings of the AAAI Conference on Artificial Intelligence},
  volume={39},
  number={22},
  pages={23378--23386},
  year={2025}
}

@article{liu2018reinforcement,
  title={Reinforcement learning on web interfaces using workflow-guided exploration},
  author={Liu, Evan Zheran and Guu, Kelvin and Pasupat, Panupong and Shi, Tianlin and Liang, Percy},
  journal={arXiv preprint arXiv:1802.08802},
  year={2018}
}

@inproceedings{shi2017world,
  title={World of bits: An open-domain platform for web-based agents},
  author={Shi, Tianlin and Karpathy, Andrej and Fan, Linxi and Hernandez, Jonathan and Liang, Percy},
  booktitle={International Conference on Machine Learning},
  pages={3135--3144},
  year={2017},
  organization={PMLR}
}

@article{hamilton1994state,
  title={State-space models},
  author={Hamilton, James D},
  journal={Handbook of econometrics},
  volume={4},
  pages={3039--3080},
  year={1994},
  publisher={Elsevier}
}

@book{sutton1998reinforcement,
  title={Reinforcement learning: An introduction},
  author={Sutton, Richard S and Barto, Andrew G and others},
  volume={1},
  number={1},
  year={1998},
  publisher={MIT press Cambridge}
}

@article{wang2024survey,
  title={A survey on large language model based autonomous agents},
  author={Wang, Lei and Ma, Chen and Feng, Xueyang and Zhang, Zeyu and Yang, Hao and Zhang, Jingsen and Chen, Zhiyuan and Tang, Jiakai and Chen, Xu and Lin, Yankai and others},
  journal={Frontiers of Computer Science},
  volume={18},
  number={6},
  pages={186345},
  year={2024},
  publisher={Springer}
}

@article{yang2023auto,
  title={Auto-gpt for online decision making: Benchmarks and additional opinions},
  author={Yang, Hui and Yue, Sifu and He, Yunzhong},
  journal={arXiv preprint arXiv:2306.02224},
  year={2023}
}

@article{shlomov2024grounding,
  title={From grounding to planning: Benchmarking bottlenecks in web agents},
  author={Shlomov, Segev and Sela, Aviad and Levy, Ido and Galanti, Liane and Abitbol, Roy and others},
  journal={arXiv preprint arXiv:2409.01927},
  year={2024}
}

@inproceedings{yao2022react,
  title={React: Synergizing reasoning and acting in language models},
  author={Yao, Shunyu and Zhao, Jeffrey and Yu, Dian and Du, Nan and Shafran, Izhak and Narasimhan, Karthik R and Cao, Yuan},
  booktitle={The eleventh international conference on learning representations},
  year={2022}
}

@article{garcez2023neurosymbolic,
  title={Neurosymbolic ai: The 3 rd wave},
  author={Garcez, Artur d’Avila and Lamb, Luis C},
  journal={Artificial Intelligence Review},
  volume={56},
  number={11},
  pages={12387--12406},
  year={2023},
  publisher={Springer}
}

@article{qiu2024llm,
  title={LLM-based agentic systems in medicine and healthcare},
  author={Qiu, Jianing and Lam, Kyle and Li, Guohao and Acharya, Amish and Wong, Tien Yin and Darzi, Ara and Yuan, Wu and Topol, Eric J},
  journal={Nature Machine Intelligence},
  volume={6},
  number={12},
  pages={1418--1420},
  year={2024},
  publisher={Nature Publishing Group UK London}
}

@article{mon2025embodied,
  title={Embodied large language models enable robots to complete complex tasks in unpredictable environments},
  author={Mon-Williams, Ruaridh and Li, Gen and Long, Ran and Du, Wenqian and Lucas, Christopher G},
  journal={Nature Machine Intelligence},
  pages={1--10},
  year={2025},
  publisher={Nature Publishing Group UK London}
}

@article{kuai2025web,
  title={Web Intelligence (WI) 3.0: in search of a better-connected world to create a future intelligent society},
  author={Kuai, Hongzhi and Huang, Jimmy X and Tao, Xiaohui and Pasi, Gabriella and Yao, Yiyu and Liu, Jiming and Zhong, Ning},
  journal={Artificial Intelligence Review},
  volume={58},
  number={9},
  pages={265},
  year={2025},
  publisher={Springer}
}

@article{abou2025agentic,
  title={Agentic AI: a comprehensive survey of architectures, applications, and future directions},
  author={Abou Ali, Mohamad and Dornaika, Fadi and Charafeddine, Jinan},
  journal={Artificial Intelligence Review},
  volume={59},
  number={1},
  pages={11},
  year={2025},
  publisher={Springer}
}

@article{lee2002intelligent,
  title={Intelligent agent-based systems for personalized recommendations in Internet commerce},
  author={Lee, Wei-Po and Liu, Chih-Hung and Lu, Cheng-Che},
  journal={Expert Systems with Applications},
  volume={22},
  number={4},
  pages={275--284},
  year={2002},
  publisher={Elsevier}
}

\appendix
\section{Prompt for The Full-Plan-in-Advance Agent} 
\label{app:promptPlanAgent}
You are an autonomous intelligent agent tasked with navigating a web browser. You will be given web-based tasks. These tasks will be accomplished through the use of specific actions you can issue. A general plan will guide your actions throughout the task.
Here's the information you'll have:
\begin{itemize}
    \item The user's objective: This is the task you're trying to complete.
    \item The current web page's accessibility tree: A simplified representation of the webpage that provides key information.
    \item The current web page's URL: The page you are currently navigating.
    \item The open tabs: The tabs that are currently open.
    \item The previous action: This is the action you have just performed. It may help you keep track of your progress.    \end{itemize}
The actions you can perform fall into several categories:

Page Operation Actions:
\begin{itemize}
    \item `click [id]`: This action clicks on an element with a specific id on the webpage.
    \item `type [id] [content] [press\_enter\_after 0  1]`: Use this to type the content into the field with id. By default, the "Enter" key is pressed after typing unless press\_enter\_after is set to 0.
    \item `hover [id]`: Hover over an element with id.
    \item `press [key\_comb]`: Simulates the pressing of a key combination on the keyboard (e.g.,Ctrl+v).
    \item `scroll [direction=down up]`: Scroll the page up or down.
    \end{itemize}

Tab Management Actions:
\begin{itemize}
    \item `new\_tab`: Open a new, empty browser tab.
    \item `tab\_focus [tab\_index]`: Switch the browser's focus to a specific tab using its index.
    \item `close\_tab`: Close the currently active tab.
\end{itemize}

URL Navigation Actions:
\begin{itemize}
    \item `goto [url]`: Navigate to a specific URL.
    \item `go\_back`: Navigate to the previously viewed page.
    \item `go\_forward`: Navigate to the next page (if a previous 'go\_back' action was performed).
\end{itemize}

Completion Action:
`stop [answer]`: Issue this action when you believe the task is complete. If the objective is to find a text-based answer, provide the answer in the bracket. If you believe the task is impossible to complete, provide the answer as "N/A" in the bracket.

To be successful, it is very important to follow the following rules: (1) You should only issue an action that is valid given the current observation. (2) You should only issue one action at a time. (3) You should follow the examples to reason step by step and then issue the next action. (4) Strictly follow the format of the actions you can perform. Start with a ``In summary, the next action I will perform is'' phrase, followed by action inside ```. For example, ``In summary, the next action I will perform is ```click [1234]```''. (5) Issue stop action when you think you have achieved the objective. Don't generate anything after stop. (6) You may use manuals to help solve the task. Documentation for GitLab, Wikipedia, an online encyclopedia and Manual on using the admin portal can be found at the homepage at http:\/\/homepage.com. (7) Follow the general plan, but always adapt to new information from observations. Look for signals that indicate failure or obstacles. If your observations suggest that the general plan may not work as expected, consider alternative actions based on the feedback you received. 

Example 1:...












Example 2:...













\section{Prompt for Generating The General Plan}
\label{app:promptGeneratePlan}

You are an intelligent agent that solves tasks on the web. Your job now is to create a GENERAL PLAN before taking any actions. Here's the information you'll have: (1) The user's objective: This is the task you're trying to complete. (2) The current web page's accessibility tree: This is a simplified representation of the webpage, providing key information. (3) The current web page's URL: This is the page you're currently navigating.

Guidelines for the general plan: (1) Include short reasoning for each step. (2) Indicate the type of action you would take (click, type, scroll, etc.). (3) Consider the page structure. Some tasks may require context about the current page. (4) Write the plan as a numbered sequence.

Return ONLY the general plan.

\section{Modifications Made to The Webarena Prompt}
\label{app:promptWebarenaAgent}
To avoid redundancy and maintain readability, the main text presents only the portions of the prompt that differ from the WebArena original prompt \citep{zhou2023webarena}. It highlights the exact modifications we introduced.

Table~\ref{tab:prompt1-compare} presents a side-by-side comparison of the original WebArena action-generation prompt and the modified version used in our work. Table~\ref{tab:promptOrderChange} presents a comparison between the original WebArena prompt structure and the reordered version used in our work. The last change that is presented in Table~\ref{tab:promptDifferentTasks} highlights two key modifications we introduce to improve the agent’s planning behavior.


\begin{table*}[h]
\centering
\caption{Comparison of the original and modified prompts. We unify the action delimiter by replacing the inconsistent five-backtick instruction with the three-backtick format used in the example.}
\label{tab:prompt1-compare}
\begin{tabular}
{|p{0.4\textwidth}| p{0.4\textwidth}|}
\hline
\textbf{Original WebArena Prompt} &
\textbf{Modified Prompt (Our Work)} \\
\hline
...To be successful, it is very important to follow the following rules:...

4. Generate the action in the correct format. Start with a ``In summary, the next action I will perform is'' phrase, followed by action inside \colorbox{yellow}{\textasciigrave\textasciigrave\textasciigrave\textasciigrave\textasciigrave}. For example, ``In summary, the next action I will perform is \verb|```|click [1234]\verb|'''|''...
 &
... To be successful, it is very important to follow the following rules:...

4. Generate the action in the correct format. Start with a ``In summary, the next action I will perform is'' phrase, followed by action inside \colorbox{yellow}{\textasciigrave\textasciigrave\textasciigrave}. For example, ``In summary, the next action I will perform is \verb|```|click [1234]\verb|'''|''....\\
\hline

\end{tabular}
\end{table*}


\begin{table*}[h]
\centering
\caption{Prompt format used in our work, showing the revised ordering of components.}
\label{tab:promptOrderChange}
\begin{tabular}
{|p{0.45\textwidth}|p{0.45\textwidth}|}
\hline
\textbf{Original WebArena Prompt} &
\textbf{Modified Prompt (Our Work)} \\
\hline
\colorbox{cyan}{OBSERVATION:}

[1744] link `HP CB782A\#ABA 640 Inkjet Fax Machine (Renewed)'

[1749] StaticText `\$279.49'

[1757] button `Add to Cart'

[1760] button `Add to Wish List'

[1761] button `Add to Compare'

\colorbox{pink}{URL:}

http://onestopmarket.com/office-products/office-electronics.html

\hl{OBJECTIVE:}

What is the price of HP Inkjet Fax Machine

\colorbox{green}{PREVIOUS ACTION:} None

 & 
\hl{OBJECTIVE:}

What is the price of HP Inkjet Fax Machine

\colorbox{cyan}{OBSERVATION:}

[1744] link `HP CB782A\#ABA 640 Inkjet Fax Machine (Renewed)'

[1749] StaticText `\$279.49'

[1757] button `Add to Cart'

[1760] button `Add to Wish List'

[1761] button `Add to Compare'

\colorbox{pink}{URL:}

http://onestopmarket.com/office-products/office-electronics.html

\colorbox{green}{PREVIOUS ACTION:} None\\
\hline
\end{tabular}

\end{table*}



\begin{table*}[h]
\centering
\caption{Side-by-side comparison of the original WebArena prompt and our modified version. The modified prompt adds previous action information and incorporates structured hints to improve multi step reasoning.}
\label{tab:promptDifferentTasks}
\begin{tabular}
{|p{0.45\textwidth}|p{0.45\textwidth}|}
\hline
\textbf{Original WebArena Prompt} &
\textbf{Modified Prompt (Our Work)} \\
\hline

\colorbox{cyan}{OBSERVATION:}

[164] textbox `Search' focused: True required: False

[171] button `Go'

[174] link `Find directions between two points'

[212] heading `Search Results'

[216] button `Close'

\colorbox{pink}{URL:}

http://openstreetmap.org

\hl{OBJECTIVE:}

\textbf{Show me the restaurants near CMU}

\colorbox{green}{PREVIOUS ACTION:} None,

Let's think step-by-step. This page has a search box whose ID is [164]. According to the nominatim rule of openstreetmap, I can search for the restaurants near a location by ``restaurants near''. I can submit my typing by pressing the Enter afterwards. In summary, the next action I will perform is \verb|```|Type [164] [restaurants near CMU] [1]\verb|'''| 
& 

\hl{OBJECTIVE:}

\textbf{Check if CMU can be reached in one hour by car from Hobart Street in Pittsburgh}

\colorbox{cyan}{OBSERVATION:}

[656] textbox `From' required: False

[649] textbox `To' required: False

[609] combobox `` hasPopup: menu expanded: False

[607] button `Go'

[629] button `Reverse Directions'

\colorbox{pink}{URL:} 

http://openstreetmap.org

\colorbox{green}{PREVIOUS ACTION:} 
click [174] where [174] is link `Find directions between two points'
						 
Let's think step-by-step. There are two textboxes available: one for the starting point and one for the destination. The one labeled `From' , which is where I will type the starting location. I will search for `Hobart Street, Pittsburgh' in the `From' textbox that has an id of [656]. After typing, I will not submit my input by pressing Enter, because I need to complete the destinetion first. ACTION: In summary, the next action I will perform is \verb|```|Type [656] [Hobart Street, Pittsburgh] [0]\verb|'''|\\
\hline
\end{tabular}
\end{table*}

\section{Prompt for The Step Success Rate Metric}
\label{app:promptStepSuccessRate}

You will be provided with two trajectories: agent steps and human gold steps, both attempting to complete the same task. 
Your task is to compare the two trajectories and identify which steps from the agent trajectory are semantically similar to the human trajectory. You must meet the following conditions for semantic similarity: Minor textual variations such as differences in casing (`Design' vs `design'), extra/missing spaces, or punctuation should be ignored as long as the core identity and meaning are preserved. Major mismatch in meaning or details (e.g., different dates).
    
You must follow these rules: (1) Each human gold step can match exactly once with an agent step. (2) If a human repeats an action, the agent must repeat it the same number of times.
Semantic Equivalence means both actions express the same intent. Your output must be a list of the actual matching agent steps, following to the rules above. 

Example 1:...




Example 2:...



        
Evaluation Steps: (1) Read both trajectories carefully. (2) Iterate through the human gold steps in order. For each human gold step find the first agent step (from the remaining unmatched steps) that is semantically equivalent to this human gold step. (3) Apply the matching rules. (4) Record matches. (5) Return a list of the matching agent steps.

\section{Prompt for Repetitiveness Rate Metric}
\label{app:promptRepetitivenessRate}

You will be given a trajectory containing a sequence of agent actions. Your goal is to identify adjacent repetitive actions.
A repetitive action is an action that repeats the meaning of the previous action, even if the wording is slightly different.
Your output must include: (1) The number of repetitive actions in the trajectory. (2) A list of the actions that were identified as repetitions (each represented as a string).

Rules: (1) Only count actions that repeat the immediately preceding action. (2) Ignore actions that are similar but do not appear right after each other. (3) A repetition is based on semantic equivalence, not exact wording.

Example 1:...



Example 2:...





Evaluation Steps: (1) Read the trajectory carefully. (2) Iterate through the steps in order and compare each action to the action immediately before it. (3) Apply the repetition rules. (4) For every repetition found, increment the counter and append the repeated action to the list.

\section{Prompt for Partial Success Rate Metric}
\label{app:promptPartialSuccessRate}

You will be given: (1) Task description. (2) A list of reference keywords/phrases that represent the essential components of a correct answer. (3) The agent’s final answer. Your goal is to check whether the agent’s final answer semantically satisfies each required keyword/phrase. Semantic equivalence means that the agent’s answer conveys the same meaning as the reference keyword/phrase, even if the wording is different. Your output should be in a JSON format:

\{"keyword1": true, "keyword2": false\}

Rules: (1) Match by meaning, not surface form. (2) Be case-insensitive.

Example:...





Evaluation Steps: (1) Read the task, the list of phrases, and the agent’s final answer carefully. (2) For each reference keyword/phrase, decide whether its meaning appears in the agent’s final answer. (3) Return a JSON object where each key is a reference keyword/phrase and each value is true/false accordingly.



\end{document}